\documentclass[onecolumn]{article}

\usepackage{arxiv}
\usepackage{amsmath,amsfonts}
\usepackage{array}
\usepackage{subfigure}
\usepackage{textcomp}
\usepackage{stfloats}
\usepackage{url}
\usepackage{verbatim}
\usepackage{graphicx}
\usepackage{cite}
\usepackage{multirow}
\usepackage[square,comma,numbers,sort&compress]{natbib}
\usepackage{booktabs}
\usepackage{caption}
\usepackage{appendix}
\usepackage{bm}
\usepackage[T1]{fontenc}
\usepackage[utf8]{inputenc}
\usepackage{hyperref}
\usepackage{url}

\usepackage{algorithm}
\usepackage{algorithmic}

\begin{document}

\title{Continual Offline Reinforcement Learning via Diffusion-based Dual Generative Replay}

\author{Jinmei Liu$~^1$\hspace{1em} Wenbin Li$~^1$\hspace{1em} Xiangyu Yue $~^2$\hspace{1em} Shilin Zhang$~^1$ \hspace{1em} Chunlin Chen$~^1$ \hspace{1em} Zhi Wang$~^1$  \\
	\normalsize $^1~$Department of Control and Systems Engineering, Nanjing University, Nanjing, China \\
	\normalsize $^2~$ Department of Information Engineering, Chinese University of Hong Kong
	\vspace{0.5em} \\
	\normalsize \tt \{jmliu, shilinzhang\}@smail.nju.edu.cn,  xyyue@ie.cuhk.edu.hk, \\ \{zhiwang, clchen, liwenbin\}@nju.edu.cn
}

\date{}

\maketitle

\begin{abstract} 
We study continual offline reinforcement learning, a practical paradigm that facilitates forward transfer and mitigates catastrophic forgetting to tackle sequential offline tasks. We propose a dual generative replay framework that retains previous knowledge by concurrent replay of generated pseudo-data. First, we decouple the continual learning policy into a diffusion-based generative behavior model and a multi-head action evaluation model, allowing the policy to inherit distributional expressivity for encompassing a progressive range of diverse behaviors. Second, we train a task-conditioned diffusion model to mimic state distributions of past tasks. Generated states are paired with corresponding responses from the behavior generator to represent old tasks with high-fidelity replayed samples. Finally, by interleaving pseudo samples with real ones of the new task, we continually update the state and behavior generators to model progressively diverse behaviors, and regularize the multi-head critic via behavior cloning to mitigate forgetting. Experiments demonstrate that our method achieves better forward transfer with less forgetting, and closely approximates the results of using previous ground-truth data due to its high-fidelity replay of the sample space.  Our code is available at \href{https://github.com/NJU-RL/CuGRO}{https://github.com/NJU-RL/CuGRO}.

\end{abstract}

\section{Introduction}
Offline reinforcement learning (RL)~\citep{fujimoto2019off,levine2020offline} allows an agent to learn from a pre-collected dataset without having to interact with the environment in real-time. 
This learning paradigm is vital for many realistic scenarios where collecting data online can be very expensive or dangerous, such as robotics~\citep{kumar2022workflow}, autonomous driving~\citep{yu2018bdd100k}, and healthcare~\citep{gottesman2019guidelines}, and has attracted widespread attention in recent years~\citep{yuan2022robust,nikulin2023anti}. 
The emergence of offline RL also holds tremendous promise for turning massive datasets into powerful sequential decision-making engines, e.g., decision transformers~\citep{chen2021decision}, akin to the rise of large language models like GPT~\citep{brown2020language}.

In the real world, huge amounts of new data are produced as new tasks emerge overwhelmingly~\citep{van2022three}.
However, current parametric RL models learn representations from stationary batches of training data, and are prone to forgetting previous knowledge when tackling new tasks, a phenomenon known as catastrophic forgetting or interference~\citep{parisi2019continual}.
Accordingly, continual RL~\citep{khetarpal2022towards}, also known as lifelong RL, is widely studied to address two major issues: i) mitigating catastrophic forgetting, and ii) allowing forward transfer, i.e., leveraging previous knowledge for efficient learning of new tasks.
In recent years, continual learning has seen the proposal of a variety of methods~\citep{fu2022model,gaya2023building} that can mainly be categorized into: regularization-based~\citep{zeng2019continual,kaplanis2019policy}, parameter isolation~\citep{kessler2022same,konishi2023parameter}, and rehearsal methods~\cite{isele2018selective,rolnick2019experience,daniels2022model}. 
Among them, rehearsal with experience replay is a popular choice due to its simplicity and promising results~\citep{gao2023ddgr}.
A recent benchmark study~\cite{wolczyk2022disentangling} demonstrates that experience replay significantly improves both transfer and the final performance, and plays a critical role in continual RL.

This paper focuses on continual offline RL (CORL), an understudied problem that lies in the intersection of offline RL and continual RL.
The learner is presented with a sequence of offline tasks where datasets are collected by some behavior policies.
Current rehearsal-based methods are confronted with two main challenges arising from intrinsic attributes of offline RL.
First, different from supervised learning, RL models are more prone to deficient generalization across diverse tasks~\citep{kirk2023survey}, and existing policy models are usually unimodal Gaussian models with limited distributional expressivity~\citep{chen2023offline}.
Nonetheless, in the realm of CORL, collected behaviors become progressively diverse as novel datasets continue to emerge, which might lead to performance degradation due to deficient generalization and distribution discrepancy.
Second, many existing methods~\citep{rolnick2019experience,wolczyk2022disentangling,gai2023oer} rely on a substantial buffer to store real samples of previous tasks.
However, this presents a memory capacity constraint that becomes more pronounced as new tasks keep emerging, restricting its applicability in large-scale problems and practical scenarios involving privacy issues.

In the paper, we propose an efficient \textbf{C}ontinual learning method via diffusion-based d\textbf{u}al \textbf{G}enerative \textbf{R}eplay for \textbf{O}ffline RL (CuGRO), which avoids storing past samples and retains previous knowledge by concurrent replay of generated pseudo-data.
First, inspired by~\cite{chen2023offline}, we decouple the continual learning policy into an expressive generative behavior model $\mu_{\bm{\phi}}(\bm{a}|\bm{s})$ and an action evaluation model $Q_{\bm{\theta}}(\bm{s},\bm{a})$.
Training a unified behavior model can continually absorb new behavior patterns to promote forward knowledge transfer for the offline setting, and sampling from this generative model can naturally encompass a progressive range of observed behaviors.
Second, we introduce a state generative model $p_{\bm{\varphi}}(\bm{s}|k)$ to mimic previous state distributions conditioned on task identity $k$. 
Generated states $\hat{\bm{s}}\sim p_{\bm{\varphi}}(\bm{s}|k)$ are paired with corresponding responses from the behavior generative model $\hat{\bm{a}}\sim\mu_{\bm{\phi}}(\bm{a}|\hat{\bm{s}})$ to represent old tasks.
In particular, we leverage existing advances in diffusion probabilistic models~\citep{ho2020denoising} to model states and corresponding behaviors with high fidelity, allowing the continual policy to inherit the distributional expressivity.
Finally, to model progressively diverse behaviors without forgetting, we interleave replayed samples with real ones of the new task to continually update the state and behavior generators.
We use a multi-head critic to tackle the diversity of emerging tasks, and mitigate forgetting of the critic in a behavior-cloning manner. 

In summary, our main contributions are threefold:
\begin{itemize}
\item We propose an efficient generative replay framework for CORL. To the best of our knowledge, CuGRO is the first that leverages expressive diffusion models to tackle the understudied CORL challenge.

\item We develop a diffusion-based dual generator system to synthesize high-fidelity samples for modeling progressively diverse behaviors, and mitigate forgetting of a multi-head critic using behavior cloning.

\item We empirically show on the MuJoCo and Meta-World benchmarks that CuGRO better mitigates forgetting and facilitates forward transfer, and closely approximates the same results as using previous ground-truth data.
\end{itemize}

\section{Preliminaries}
\subsection{Offline Reinforcement Learning}
RL is commonly studied based on the Markov decision process (MDP) formulation, $(\mathcal{S}, \mathcal{A}, P, r, \gamma)$, where $\mathcal{S}$ and $\mathcal{A}$ denote the state and action spaces, $P(\bm{s}'|\bm{s},\bm{a})$ and $r(\bm{s},\bm{a})$ are the transition and reward functions, and $\gamma$ is the discount factor.
The goal is to maximize the expected return:
\begin{equation}
\begin{aligned}
J(\pi) =\int_{\mathcal{S}}\rho_{\pi}(\bm{s})\int_{\mathcal{A}}\pi(\bm{a}|\bm{s})Q^{\pi}(\bm{s},\bm{a})\mathrm{d}\bm{a}\mathrm{d}\bm{s},
\end{aligned}
\end{equation}
where $\rho_{\pi}(\bm{s})$ is the state visitation frequencies induced by policy $\pi$, $Q^{\pi}(s,a)=\mathbb{E}_{\pi}[\sum_{t'=0}^{\infty}\gamma^{t'}r_{t+t'}|s_t=s,a_t=a]$ is the expected future rewards (return-to-go).
When online data collection from $\pi$ is infeasible, it is difficult to estimate $\rho_{\pi}(\bm{s})$ and thus $J(\pi)$.
For a static dataset $\mathcal{D}_{\mu}=\sum_i(\bm{s}_i,\bm{a}_i,r_i,\bm{s}_i')$ collected by a behavior policy $\mu(\bm{a}|\bm{s})$, offline RL approaches~\citep{peng2019advantage,nair2020awac} usually encourage $\pi$ to stick with $\mu$ and maximize a constrained objective as
\begin{equation}
\begin{aligned}
J'(\pi) =  \int_{\mathcal{S}}\rho_{\mu}(\bm{s})\int_{\mathcal{A}}\pi(\bm{a}|\bm{s})Q^{\pi}(\bm{s},\bm{a})\mathrm{d}\bm{a}\mathrm{d}\bm{s} 
- \frac{1}{\alpha}\int_{\mathcal{S}}\rho_{\mu}(\bm{s})D_{\text{KL}}\left(\pi(\cdot|\bm{s}) || \mu(\cdot|\bm{s})\right) \mathrm{d}\bm{s}.
\label{obj}
\end{aligned}
\end{equation}
The optimal policy $\pi^*$ for Eq.~(\ref{obj}) can be derived by using a Lagrange multiplier as 
\begin{equation}
\pi^*(\bm{a}|\bm{s}) = \frac{1}{Z(\bm{s})}\mu(\bm{a}|\bm{s})\exp\left( \alpha Q^*(\bm{s}, \bm{a}) \right),
\end{equation}
where $Z(\bm{s})$ is the partition function.
By projecting $\pi^*$ onto a parameterized policy $\pi_{\bm{\phi}}$ with a critic $Q_{\bm{\theta}}$, we can obtain the final objective in a weighted regression form as 
\begin{equation}
\begin{aligned}
\underset{{\bm{\phi}}}{\arg\min}~\mathbb{E}_{\bm{s}\sim\mathcal{D}_{\mu}}\left[ D_{\text{KL}}\left( \pi^* || \pi_{\bm{\phi}} \right) \right] = 
\underset{{\bm{\phi}}}{\arg\max}~\mathbb{E}_{(\bm{s},\bm{a})\sim\mathcal{D}_{\mu}}\left[ \frac{\log\pi_{\bm{\phi}}(\bm{a}|\bm{s})\exp\left( \alpha Q_{\bm{\theta}}(\bm{s}, \bm{a}) \right)}{Z(\bm{s})} \right].
\end{aligned}
\label{obj2}
\end{equation}

\subsection{Generative Behavior Modeling}
To avoid learning an explicitly parameterized policy model in Eq.~(\ref{obj2}), ~\citep{chen2023offline} decouples the learned policy into an expressive generative behavior model $\mu_{\bm{\phi}}(\bm{a}|\bm{s})$ and an action evaluation model $Q_{\bm{\theta}}(\bm{s},\bm{a})$, and form a policy improvement step as
\begin{equation}
\pi(\bm{a} | \bm{s}) \propto \mu_{\bm{\phi}}(\bm{a} | \bm{s}) \cdot \exp\left(\alpha Q_{\bm{\theta}}(\bm{s}, \bm{a})\right),
\label{decouple}
\end{equation}
where temperature $\alpha$ balances between conservative and greedy improvements. Diffusion probabilistic models~\citep{ho2020denoising,song2021score} are utilized to fit the behavior distribution from the offline dataset.
A state-conditioned diffusion model $\bm{\epsilon}_{\bm{\phi}}$ is trained to predict the noise $\bm{\epsilon}$ added to the action $\bm{a}$ sampled from the behavior policy $\mu(\cdot|\bm{s})$ as
\begin{equation}
\underset{\bm{\phi}}{\arg \min}~\mathbb{E}_{(\bm{s}, \bm{a}) \sim D_\mu, \boldsymbol{\epsilon}, t}\left[\left\|\sigma_t \bm{\epsilon}_{\bm{\phi}}\left(\alpha_t \boldsymbol{a}+\sigma_t \boldsymbol{\epsilon}, \boldsymbol{s}, t\right)+\boldsymbol{\epsilon}\right\|_2^2\right],
\label{action_update}
\end{equation}
where $t \sim \mathcal{U}(0, T)$, $\boldsymbol{\epsilon} \sim \mathcal{N}(0, \boldsymbol{I})$, and $\alpha_t$ and $\sigma_t$ are determined by the forward diffusion process. 
The generative model $\bm{\epsilon}_{\bm{\phi}}$ is trained to denoise the perturbed action $\bm{a}_t:=\alpha_t \boldsymbol{a}+\sigma_t \boldsymbol{\epsilon}$ back to the original undisturbed one $\boldsymbol{a}$.
Then, a critic network $Q_{\bm{\theta}}(\bm{s},\bm{a})$ is learned to evaluate actions sampled from the learned behavior model $\mu_{\bm{\phi}}(\cdot|\bm{s})$.
With the idea of episodic learning~\citep{blundell2016model,ma2022offline}, the value function is updated using a planning-based Bellman operator as
\begin{eqnarray}
\!\!\!\!&&\!\!\!\! R_i^{(j)} = r_i + \gamma\max\left( R_{i+1}^{(j)},  V_{i+1}^{(j-1)}\right), \nonumber \\
\text{where} \!\!\!\!&&\!\!\!\! V_i^{(j-1)}:=\mathbb{E}_{\bm{a}\sim\pi(\cdot|\bm{s}_i)} \left[ Q_{\bm{\theta}}(\bm{s}_i, \bm{a}) \right], \label{qlearning}\\
\text{and} \!\!\!\!&&\!\!\!\! \bm{\theta} \!=\! \underset{\bm{\theta}}{\arg\min}~ \mathbb{E}_{(\bm{s}_i, \bm{a}_i)\sim \mathcal{D}_{\mu}} \!\!\left[\Vert Q_{\bm{\theta}}(\bm{s}_i, \bm{a}_i) \!-\! R_i^{(j-1)} \Vert_2^2 \right]\!\!, \nonumber 
\end{eqnarray}

where $j=1,2,...$ is the iteration number, $R_i^{(0)}=\sum_{i'=0}^{\infty}r_{i+i'}$ is the vanilla return of trajectories.
Eq.~(\ref{qlearning}) offers an implicit planning scheme within dataset trajectories to avoid bootstrapping over unseen actions and to accelerate convergence.
Finally, the action is re-sampled using importance weighting with $\exp(\alpha Q_{\bm{\theta}}(\bm{s},\bm{a}))$ being the sampling weights as in Eq.~(\ref{decouple}).

\textbf{Notation}. 
There are three ``times" at play in this work: that of the RL problem, that of the diffusion process, and that of the continual learning setting. 
We use subscripts $i$ to denote the timestep within an MDP, superscripts $t$ to denote the diffusion timestep, and subscripts $k$ to denote the sequential task identity.
For example, $\bm{a}_i^0$ refers to the $i$-th action in a noiseless trajectory, and $\bm{s}_k$ and $\bm{\phi}_k$ represent a given state and the behavior model during task $M_k$.

\begin{figure}[!h]\centering
\subfigure[Training the state generator.]{\includegraphics[width=0.265\textwidth]{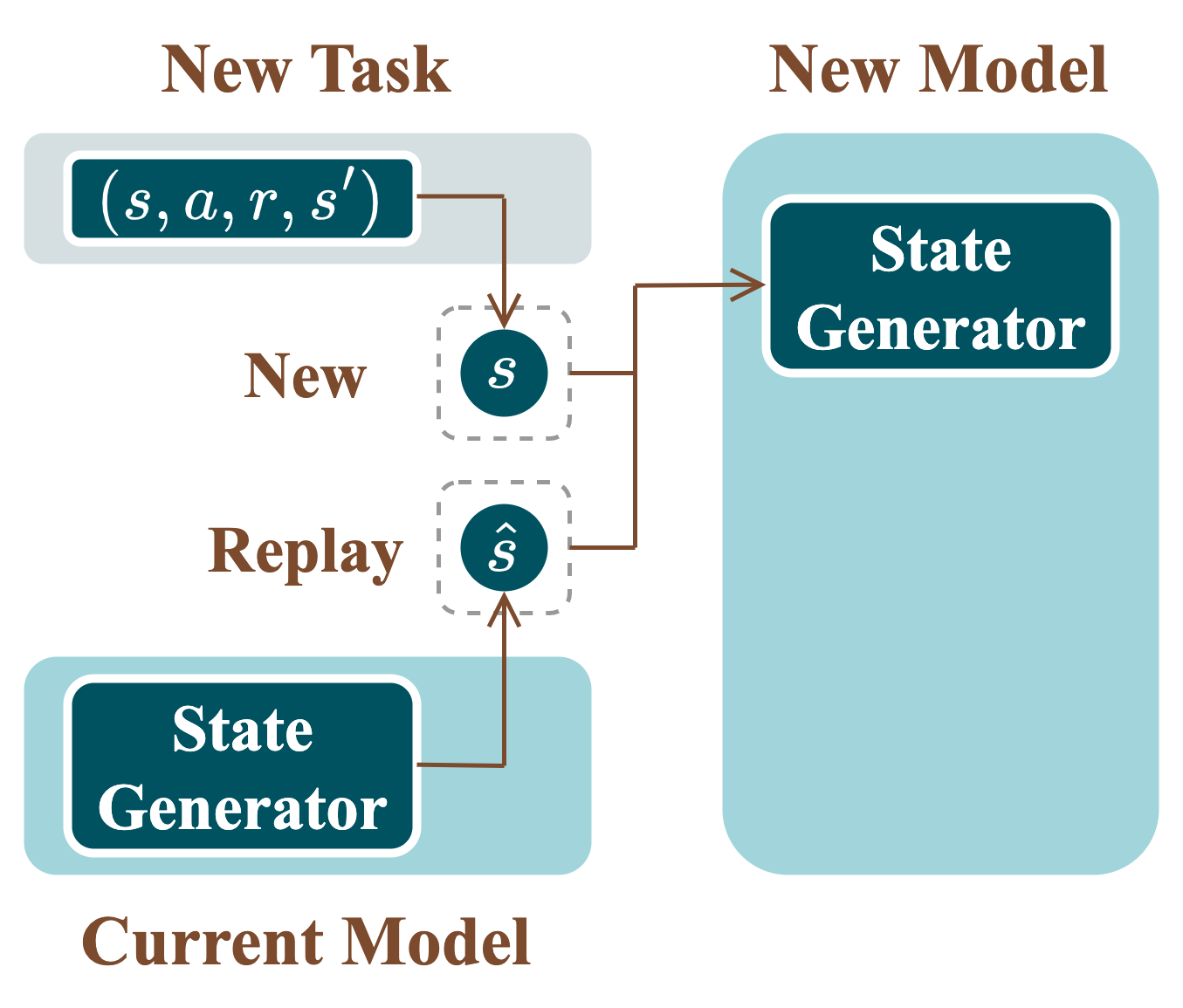}}
\subfigure[Training the behavior generator.]{\includegraphics[width=0.383\textwidth]{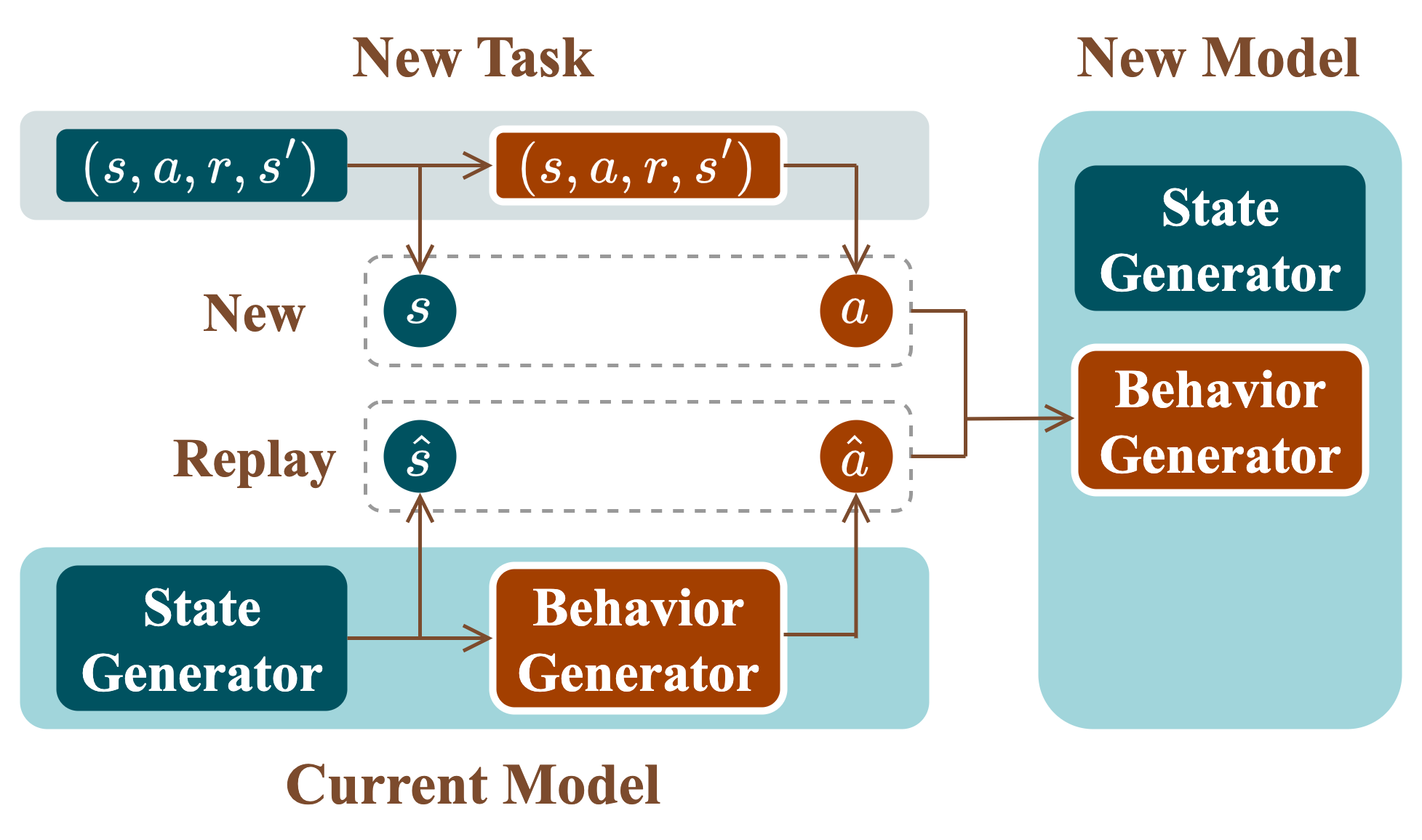}}
\subfigure[Training the multi-head critic.]{\includegraphics[width=0.5\textwidth]{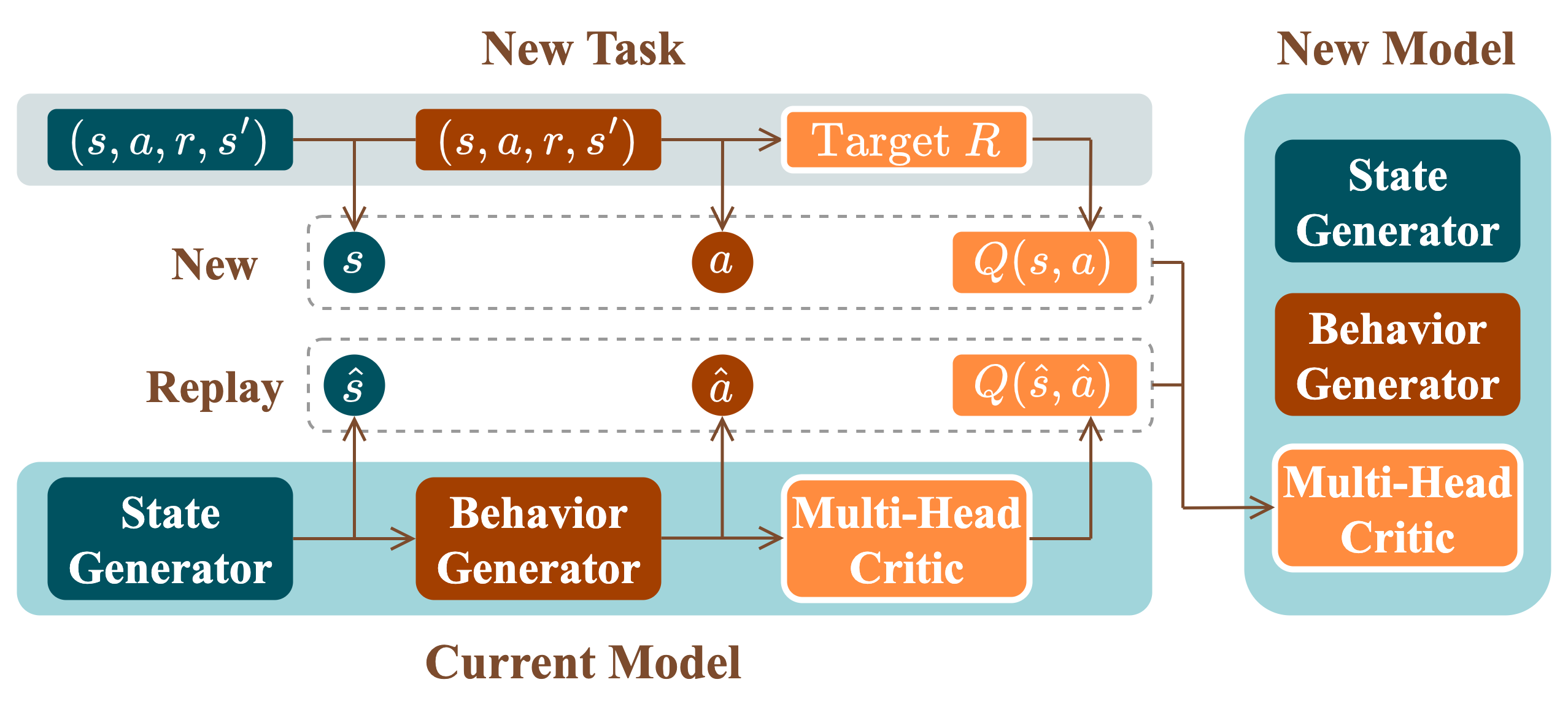}}
\caption{Sequential training of CuGRO. 
(a) A new diffusion-based state generative model is trained to mimic a mixed data distribution of real samples $\bm{s}$ and replayed ones $\hat{\bm{s}}$.
(b) A new diffusion-based behavior generative model learns from real state-action pairs $(\bm{s},\bm{a})$ and pseudo pairs $(\hat{\bm{s}}, \hat{\bm{a}})$, where replayed action $\hat{\bm{a}}$ is obtained by feeding replayed states $\hat{\bm{s}}$ into current behavior generator.
(c) A new head in the critic is expanded for tackling the new task with real state-action pairs and Bellman targets, and previous heads are regularized by cloning the Q-value of replayed pairs $(\hat{\bm{s}}, \hat{\bm{a}})$.
}
\label{method_fig}
\end{figure}

\section{Diffusion-based Dual Generative Replay}

\subsection{Problem Formulation and Method Overview}
In CORL, we assume that the task follows a distribution $M_k=(\mathcal{S},\mathcal{A},P,r,\gamma)_k\sim P(M)$.
The learner is presented with an infinite sequence of tasks $[M_1,...,M_k,...]$, and for each task $M_k$, an offline dataset $\mathcal{D}_{\mu_k}=\sum_i(\bm{s}_i,\bm{a}_i,r_i,\bm{s}_i')_k$ is collected by a behavior policy $\mu_k$.
The learner can only access the offline dataset of the new task $M_K$, without retrieving real samples from previous tasks $\{M_k\}_{k=1}^{K-1}$ or any online environmental interactions.
The objective for CORL is to learn a continual policy that maximizes the expected return over all encountered tasks as
\begin{equation}
J(\pi_{\text{continual}}) = \sum\nolimits_{k=1}^K J_{M_k}(\pi_{M_k}).
\end{equation}
Importantly, the learner ought to build upon the accumulated knowledge from previous tasks $M_1,...,M_{K-1}$ to facilitate learning the new task $M_K$, while not forgetting previously acquired knowledge.

In Section~\ref{system}, we first introduce the dual generator system that mimics the state and behavior distributions of past tasks.
Then, we present the dual generative replay mechanism in Section~\ref{dual_gr} that sequentially trains conditional diffusion-based generators to synthesize high-fidelity pseudo samples for modeling progressively diverse behaviors.
Finally, we show the behavior cloning process that prevents forgetting of a multi-head critic in Section~\ref{bc}.
With the above implementations, the training procedure is illustrated in Fig.~\ref{method_fig}, and the algorithm summary is presented in Appendix~A.

\subsection{Dual Generator System}\label{system}
\textbf{Behavior Generative Model}.
Inspired by~\cite{chen2023offline}, we decouple the continual learning policy into a diffusion-based generative behavior model $\mu_{\bm{\phi}}(\bm{a}|\bm{s})$ and an action evaluation model $Q_{\bm{\theta}}(\bm{s},\bm{a})$ as in Eq.~(\ref{decouple}).
This perspective rooted in generative modeling presents three promising advantages for CORL.
First, existing policy models are usually unimodal Gaussians with limited distributional expressivity, while collected behaviors in CORL become progressively diverse as novel datasets keep emerging.\footnote{For instance, given a bidirectional navigation task where the agent can go to either of two opposite directions to get the final reward, fitting the policy with an unimodal Gaussian distribution leads to encompassing the low-density area situated between the two peaks~\citep{chen2023offline}.}
Learning a generative behavior model is considerably simpler since sampling from the behavior policy can naturally encompass a diverse range of observed behaviors, and allows the policy to inherit the distributional expressivity of diffusion models.
Second, RL models are more prone to deficient generalization across diverse tasks~\citep{kirk2023survey}.\footnote{Given two navigation tasks where the goals are in adverse directions, the single RL model ought to execute completely opposite decisions under the same states for the two tasks. By contrast, a supervised classification model just needs to derive two decision boundaries~\citep{li2017learning} or a merged one~\citep{shin2017continual} based on the augmented feature space shared by two tasks.}
Learning a unified behavior model can naturally absorb novel behavior patterns, continually promoting knowledge transfer and generalization for offline RL.
Third, generative behavior modeling can harness extensive offline datasets from a wide range of tasks with pretraining, serving as a foundation model to facilitate finetuning for any downstream tasks.
This aligns with the paradigm of large language models~\citep{brown2020language}, and we reserve this promising avenue for future research.

\textbf{State Generative Model}.
Under the generative replay framework, we train another diffusion-based generator to mimic state distributions of previous tasks $\{\bm{s}_k\}_{k=1}^{K-1}$.
Specifically, we learn a scored-based task-conditioned model $\bm{\epsilon}_{\bm{\varphi}}$ to predict the noise $\bm{\epsilon}$ added to the state $\bm{s}$ as 
\begin{equation}
\arg\min_{\bm{\varphi}}\mathbb{E}_{\bm{s} \sim \mathcal{D}_{\mu_k}}\left[\left\|\sigma_t \bm{\epsilon}_{\bm{\varphi}}\left(\alpha_t \boldsymbol{s}+\sigma_t \boldsymbol{\epsilon}, k, t\right)+\boldsymbol{\epsilon}\right\|_2^2\right],
\label{noise_state}
\end{equation}
where $t \sim \mathcal{U}(0, T)$, $\boldsymbol{\epsilon} \sim \mathcal{N}(0, \boldsymbol{I})$, and $\alpha_t$ and $\sigma_t$ are hyperparameters of the forward diffusion process, and state $\bm{s}$ is sampled from the $k$-th task's dataset $\mathcal{D}_{\mu_k}$. 
The generative model $\bm{\epsilon}_{\bm{\varphi}}$ is trained to refine the perturbed state $\bm{s}^t:=\alpha_t \bm{s}+\sigma_t \bm{\epsilon}$ back to the original undisturbed one $\bm{s}$, such that the random noise $\boldsymbol{s}^T \sim \mathcal{N}(0, \bm{I})$ can be reversed to model the original state distribution.

To generate high-fidelity state samples for each past task, we condition the diffusion model on task identity $k$ as $\hat{\bm{s}} \sim p_{\bm{\varphi}}(\bm{s}|k)$.
Moreover, conditional diffusion models can further improve the sample quality with classifier guidance that trades off diversity for fidelity using gradients from a trained classifier~\citep{dhariwal2021diffusion}. 
Unfortunately, in CORL, it might be infeasible to train a classifier with task identity as the label, since new tasks continually emerge and samples from prior tasks are unavailable.
We leave this line of model improvement as future work, such as training with classifier-free guidance~\citep{ho2022classifier}.

\subsection{High-Fidelity Dual Generative Replay}\label{dual_gr}
We consider sequential training on dual generative models $\{(\bm{\varphi}_k, \bm{\phi}_k)\}_{k=1}^K$, where the $K$-th models $(\bm{\varphi}_{K}, \bm{\phi}_{K})$ learn the new task $M_K$ and the knowledge of previous models $(\bm{\varphi}_{K-1}, \bm{\phi}_{K-1})$.
This involves a dual procedure of training the conditional diffusion-based state and behavior generators using the mechanism of generative replay.

\textbf{State Generative Replay.}
At new task $M_K$, to prevent forgetting of the state generative model, we use the previously learned generator $\bm{\varphi}_{K-1}$ to generate synthetic state samples for retaining knowledge of prior tasks, i.e., $\hat{\bm{s}}_k\sim p_{\bm{\varphi}_{K-1}}(\bm{s}|k)$.
The new state generator $\bm{\varphi}_K$ receives real state samples of the new task $\bm{s}_K$ and replayed state samples of all previous tasks $\{\hat{\bm{s}}_k\}_{k=1}^{K-1}$.
Real and replayed samples are mixed at a ratio that depends on the desired importance of the new task compared to the older ones.
The state generator learns to mimic a mixed data distribution of real samples and replayed ones from the previous generator, aiming to reconstruct the cumulative state space.
Formally, the loss function of the $K$-th state generative model is:
\begin{equation}
\begin{aligned}
& \mathcal{L}(\bm{\varphi}_K) =
 \mathbb{E}_{\bm{s}_K\sim\mathcal{D}_{\mu_K}}\left[\left\|\sigma_t \bm{\epsilon}_{\bm{\varphi}_K}\left(\alpha_t \boldsymbol{s}_K + \sigma_t \boldsymbol{\epsilon}, K, t\right) + \boldsymbol{\epsilon}\right\|_2^2\right] + \\
& \beta\sum_{k=1}^{K-1} \mathbb{E}_{\hat{\bm{s}}_k\sim p_{\bm{\varphi}_{K-1}}(\bm{s}|k)}\left[\left\|\sigma_t \bm{\epsilon}_{\bm{\varphi}_K}\left(\alpha_t \hat{\bm{s}}_k + \sigma_t \boldsymbol{\epsilon}, k, t\right) + \boldsymbol{\epsilon}\right\|_2^2\right],
\end{aligned}
\label{train_state}
\end{equation}
where $\beta$ is the ratio of mixing replayed data.

\textbf{Behavior Generative Replay.}
Subsequently, it is crucial to replay samples from previous tasks to continually train the behavior generative model $\mu_{\bm{\phi}}$ for encompassing an expanding array of behavior patterns without forgetting.
This update involves generating pseudo state-action pairs using both generators.
At the new task $M_K$, we first obtain synthetic state samples of previous tasks using the learned state generator $\bm{\varphi}_{K-1}$, and then pair these pseudo state samples with corresponding responses from the current behavior generator $\bm{\phi}_{K-1}$ to represent old tasks as
\begin{equation}
\begin{aligned}
\hat{\bm{s}}_k \sim p_{\bm{\varphi}_{K-1}}(\bm{s}|k), 
\hat{\bm{a}}_k \sim \mu_{\bm{\phi}_{K-1}}(\bm{a}|\hat{\bm{s}}_k),~~~k=1,...,K-1.
\end{aligned}
\label{pseudo_sa}
\end{equation}
The new behavior generator $\bm{\phi}_K$ receives real state-action pairs of the new task $(\bm{s}_K, \bm{a}_K)$ and replayed pairs of all previous tasks $\{(\hat{\bm{s}}_k, \hat{\bm{a}}_k)\}_{k=1}^{K-1}$.
The behavior generator aims to continually reconstruct the ever-expanding behavior patterns.
With replayed pairs generated by Eq.~(\ref{pseudo_sa}), the loss function of the $K$-th behavior generative model is:

\begin{eqnarray}
&& \mathcal{L}(\bm{\phi}_K) = \mathbb{E}_{(\bm{s}_K,\bm{a}_K)\sim\mathcal{D}_{\mu_K}}\left[\left\|\sigma_t \bm{\epsilon}_{\bm{\phi}_K}\left(\alpha_t \boldsymbol{a}_K + \sigma_t \boldsymbol{\epsilon}, \bm{s}_K, t\right) + \boldsymbol{\epsilon}\right\|_2^2\right] \nonumber\\
&& + \beta \sum_{k=1}^{K-1} \mathbb{E}_{(\hat{\bm{s}}_k, \hat{\bm{a}}_k)}\left[\left\|\sigma_t \bm{\epsilon}_{\bm{\phi}_K}\left(\alpha_t \hat{\bm{a}}_k + \sigma_t \boldsymbol{\epsilon}, \hat{\bm{s}}_k, t\right) + \boldsymbol{\epsilon}\right\|_2^2\right].
\label{train_behavior}
\end{eqnarray}

\subsection{Behavior Cloning}\label{bc}
To tackle the diversity of emerging tasks, we employ a multi-head critic network $\bm{\theta}$, with each head $Q_{\bm{\theta}}^k(\bm{s},\bm{a})$ for capturing distinctive characteristics of each task $M_k$.
When training the new task $M_{K}$, we use the planning-based operator in Eq.~(\ref{qlearning}) to update the $K$-th Q-function $Q_{\bm{\theta}}^K(\bm{s},\bm{a})$ with real samples from newest offline dataset $(\bm{s}_K,\bm{a}_K)\sim\mathcal{D}_{\mu_K}$.
Then, we utilize a behavior cloning technique to alleviate forgetting of previous tasks' heads in the critic, analogous to the data rehearsal approaches that work well in supervised continual learning~\citep{li2017learning,chaudhry2019tiny}.
After obtaining the pseudo state-action pairs of previous tasks $\{(\hat{\bm{s}}_k, \hat{\bm{a}}_k)\}_{k=1}^{K-1}$ in Eq.~(\ref{pseudo_sa}), we annotate these replayed samples using the previously trained critic $\bm{\theta}_{K-1}$ as $\{Q_{\bm{\theta}_{K-1}}^k(\hat{\bm{s}}_k, \hat{\bm{a}}_k)\}_{k=1}^{K-1}$.
During training the new critic $\bm{\theta}_{K}$, we treat the labeled pseudo samples as expert data and perform behavior cloning via applying an auxiliary regularization term to imitate the expert data. 
The overall loss function for the multi-head critic is formulated as
\begin{eqnarray}
\quad \quad \mathcal{L}(\bm{\theta}_K) = \mathbb{E}_{(\bm{s}_K,\bm{a}_K)\sim\mathcal{D}_{\mu_K}}\left[ || Q_{\bm{\theta}_K}^K(\bm{s}_K,\bm{a}_K)- R_K||_2^2 \right] + \nonumber
\lambda\sum_{k=1}^{K-1}\mathbb{E}_{(\hat{\bm{s}}_k, \hat{\bm{a}}_k)}\left[\left(Q_{\bm{\theta}_K}^k(\hat{\bm{s}}_k, \hat{\bm{a}}_k) - Q_{\bm{\theta}_{K-1}}^k(\hat{\bm{s}}_k, \hat{\bm{a}}_k)  \right)^2 \right],
\label{train_critic}
\end{eqnarray}
where $R_K$ corresponds to the Bellman target in Eq.~(\ref{qlearning}), and $\lambda$ is the regularization coefficient.
The first term is responsible for incorporating new knowledge using an expanded head in the critic, and the second term encourages the existing heads to stay close to previous outputs for mitigating forgetting.
The critic, equipped with progressive heads, accumulates knowledge of previous tasks, enabling the reuse of the most relevant parts from the past to enhance the training of new tasks.
Additionally, the behavior cloning loss acts as a regularizer that helps shape more general features, thus further improving forward knowledge transfer.

\section{Experiments}
We evaluate CuGRO's applicability on two classical benchmarks: the multi-task MuJoCo control~\citep{todorov2012mujoco,ni2023metadiffuser} and Meta-World~\citep{yu2020meta}. 
Generative replay is superior to other continual learning methods in that the generator is the only constraint of task performance.
When the generative model is optimal, training the networks with generative replay is equivalent to joint training on the entire dataset.
This is also a significant motivation for our method to leverage powerful diffusion models as our generators, as diffusion models have achieved sample quality superior to the current state-of-the-art generative models~\citep{dhariwal2021diffusion}.

In general, we aim to empirically answer the following questions:
i) Can CuGRO model progressively diverse behaviors to achieve performance gain on CORL compared to various baselines? (Sec.~\ref{exp:main})
ii) Can CuGRO accurately mimic the mixed data distribution to produce efficient generative replay? (Sec.~\ref{exp:ablation})
iii) Can CuGRO be a superior generative framework compared to using other generative models like VAEs and GANs? (Sec.~\ref{exp:difusion})

\textbf{Datasets.} 
In the MuJoCo testbed, we consider three representative domains: \textit{Swimmer-Dir}, \textit{Walker2D-Params}, and \textit{HalfCheetah-Vel}, where each continual learning unit contains a sequence of five tasks.
In the Meta-World testbed, we select a collection of five robot manipulation tasks to conduct continual learning.
We use SAC~\citep{haarnoja2018soft} or TD3~\citep{fujimoto2018addressing} to train a single-task policy independently for each task, and we collect datasets of three different qualities. More details are provided in Appendix~B.
We primarily present the experimental results of one of the offline datasets, which was constructed with the complete buffer preserved during training. For additional experimental results, please refer to Appendix~G.

\textbf{Baselines.} 
We compare CuGRO to five competitive baselines that cover the three major categories of continual learning methods: the rehearsal-based 1) \textit{ClonEx}~\citep{wolczyk2022disentangling}, 2) \textit{Behavioral Cloning (BC)}~\citep{isele2018selective}, and 3) \textit{LwF}~\citep{li2017learning}, 4) the parameter isolation-based \textit{PackNet}~\citep{mallya2018packnet}, and 5) the regularization-based \textit{EWC}~\citep{kirkpatrick2017overcoming}.
As existing continual RL methods are usually explored within online settings, we modify these baselines to offline settings for a fair comparison and implement them based on the classical offline RL algorithm AWAC~\citep{nair2020awac}.
More details are given in Appendix~C.

All methods are carried out with $10$ different random seeds, and the mean of the received return is plotted with 95\% bootstrapped confidence intervals of the mean (shaded).
The standard errors are presented for numerical results.
Implementation details of CuGRO are given in Appendix~D.

\subsection{Performance on MuJoCo and Meta-World}\label{exp:main}
\begin{figure*}[tb]\centering
 \subfigure[Swimmer-Dir]{\includegraphics[width=0.25\textwidth]{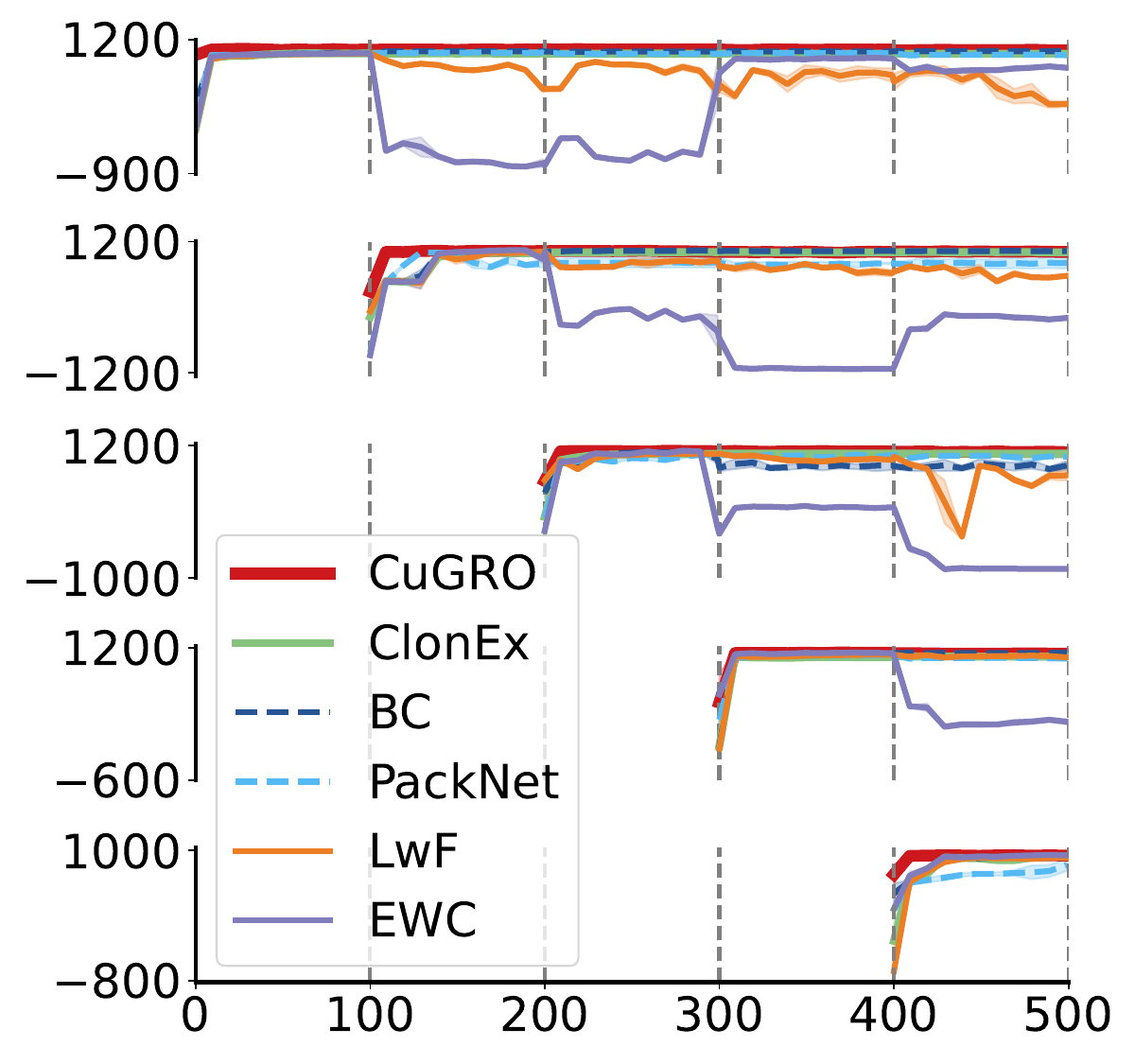}}%
 \subfigure[Walker2D-Params]{\includegraphics[width=0.25\textwidth]{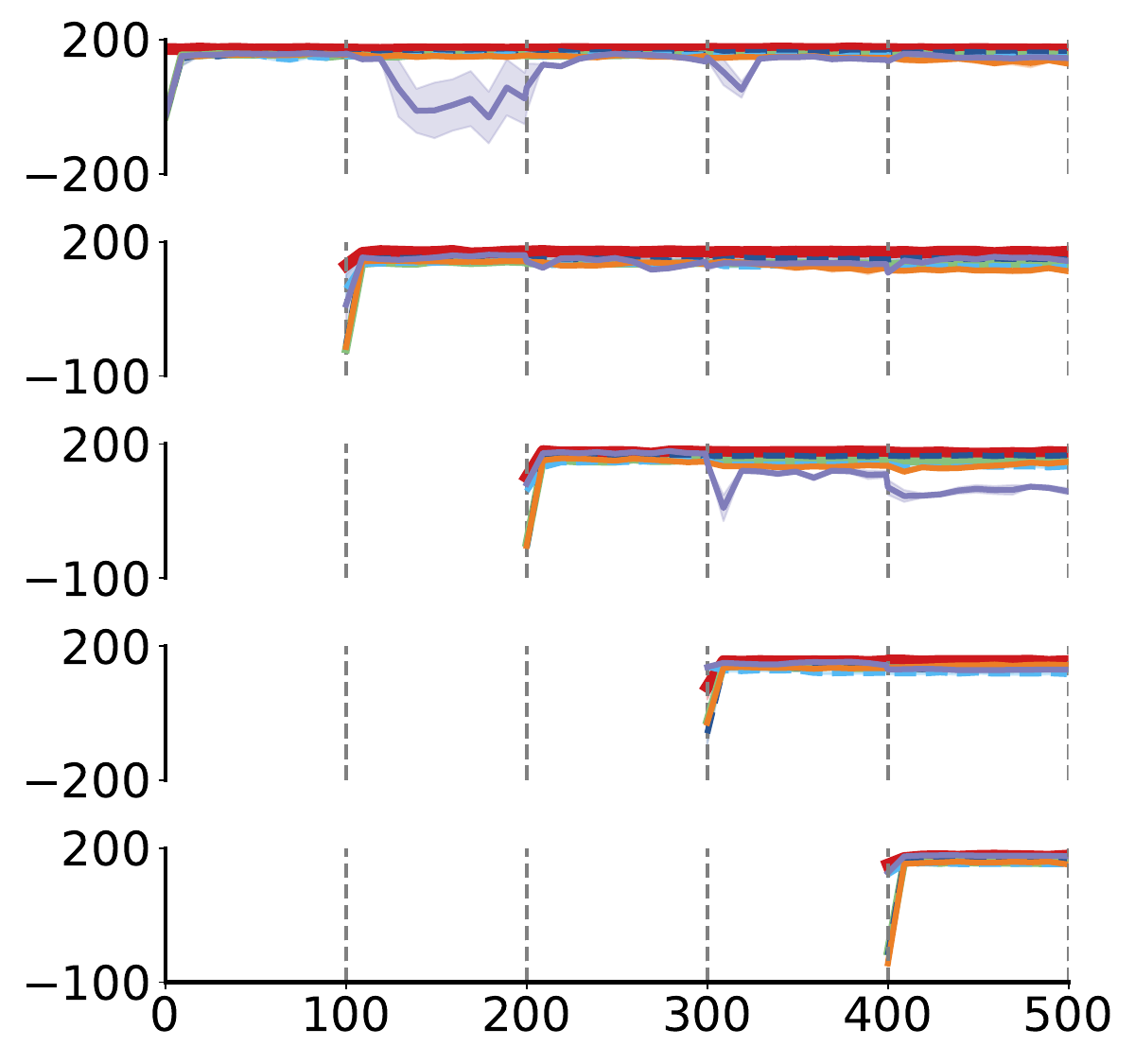}}%
 \subfigure[HalfCheetah-Vel]{\includegraphics[width=0.25\textwidth]{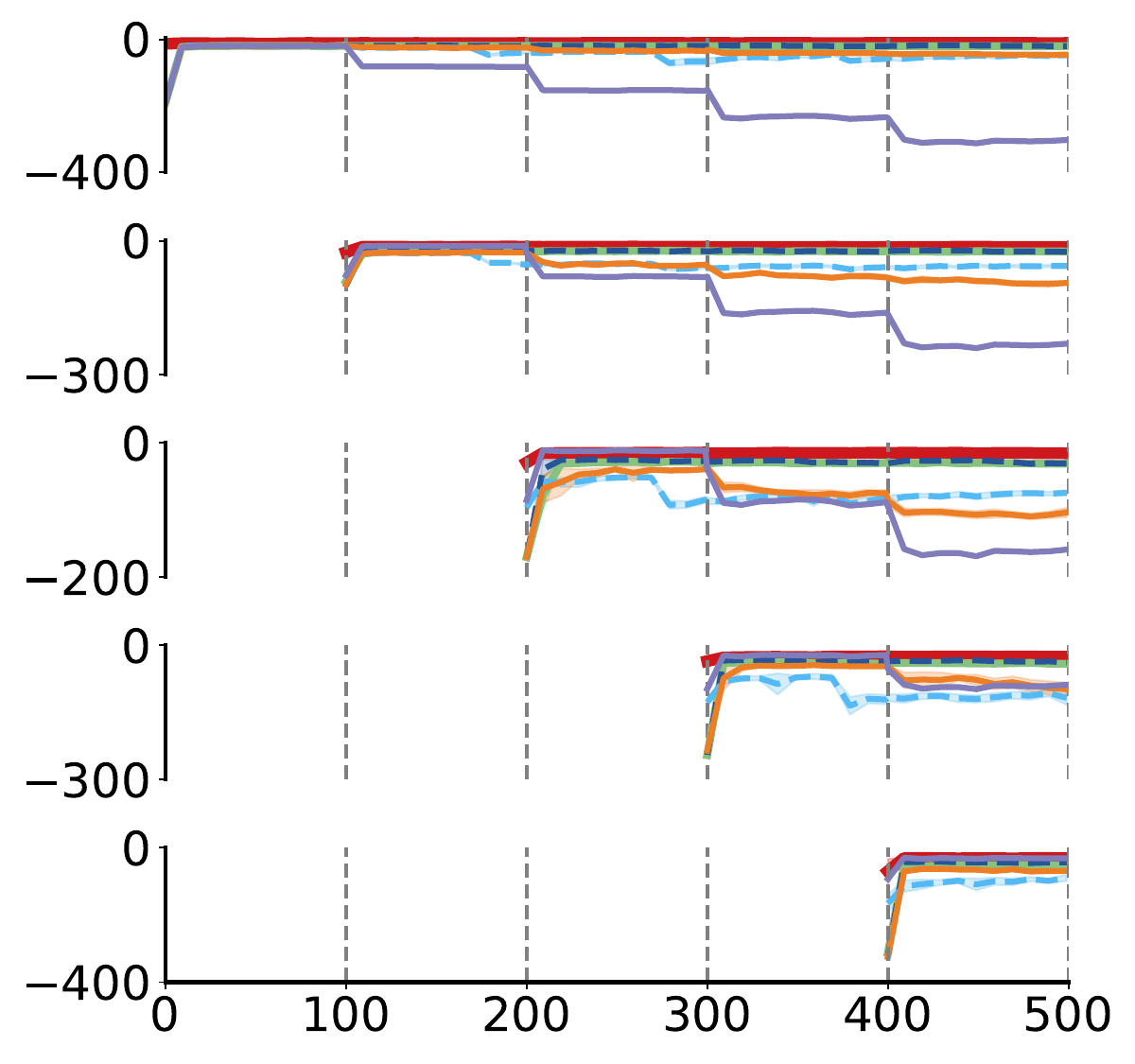}}%
\subfigure[Meta-World]{\includegraphics[width=0.25\textwidth]{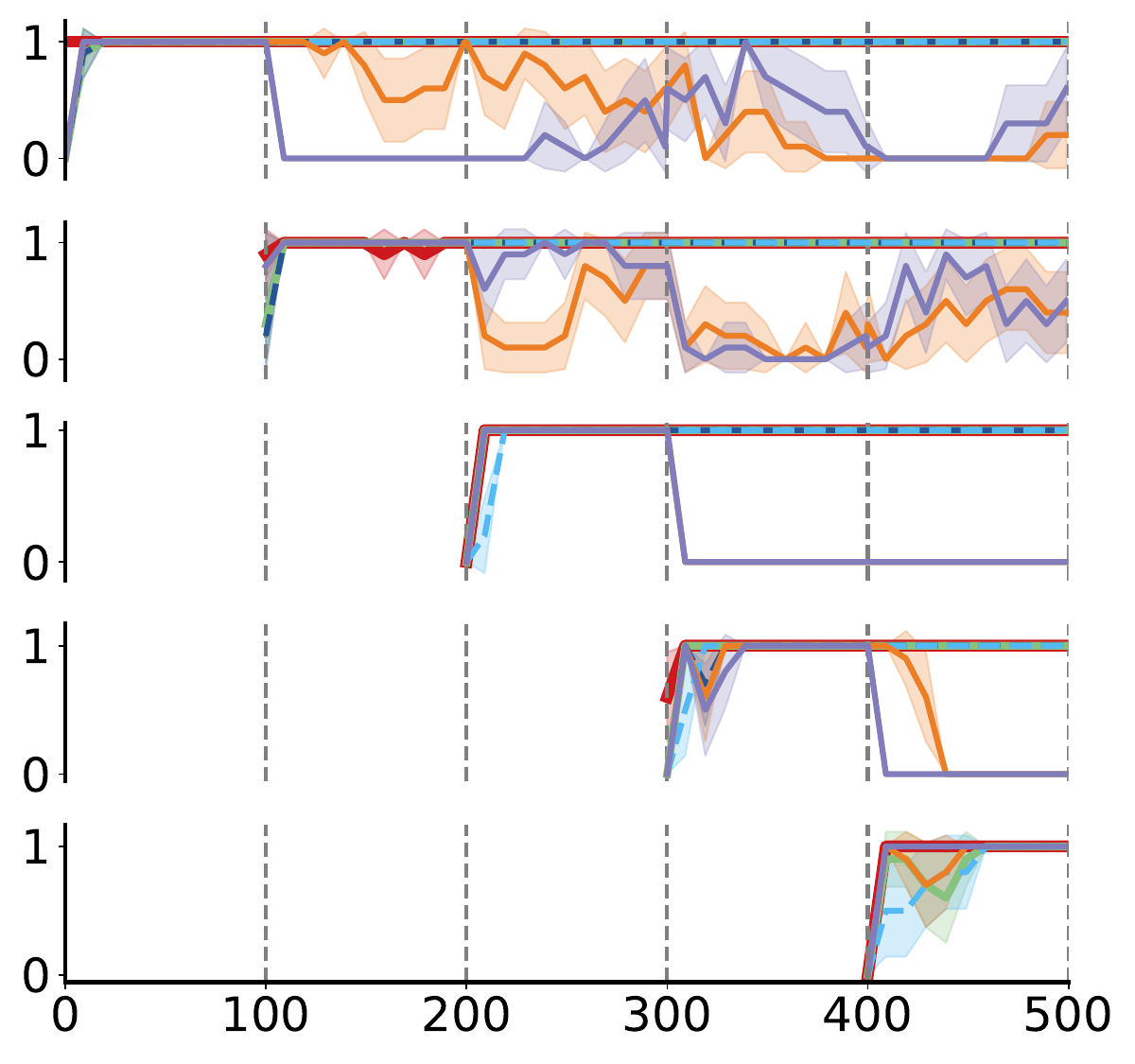}}
 \caption{Performance of CuGRO and baselines on cumulative tasks during sequential training evaluated on MuJoCo and Meta-World. The policy is trained with $100$ iterations for each task, and each iteration involves feeding all data batches to the policy model for training.}
\label{fig:results_all} 
\end{figure*}

Fig.~\ref{fig:results_all} shows the sequential training performance of CuGRO and all baselines measured on cumulative tasks, and Table~\ref{tab:results_all} presents the corresponding final performance averaged over all sequential tasks.
It can be observed that CuGRO obtains superior continual learning performance on both testbeds.
During sequential training, CuGRO quickly adapts to the new task using only a few iterations, while incurring almost zero performance loss on previous tasks.
Three baselines of ClonEx, BC, and PackNet achieve good performance for CORL tasks.
Note that ClonEx and BC exploit real samples of previous tasks for rehearsal of past knowledge, which might be expensive and even infeasible for practical scenarios.
LwF obtains worse performance with a little forgetting of previous knowledge, which could result from distribution bias since it assumes the same input distribution for all tasks.
EWC obtains poor continual learning performance with significant catastrophic forgetting, indicating its difficulty in balancing previous and new tasks with limited neural architectural resources.

\begin{table*}[!h]
\centering
\setlength{\tabcolsep}{5mm}
\caption{Final performance of CuGRO and baselines averaged over all sequential tasks evaluated on MuJoCo and Meta-World.}
\renewcommand\arraystretch{1}
\begin{tabular}{c|c|c|c|c}
\cmidrule[\heavyrulewidth]{1-5}
Method  & Swimmer-Dir  & Walker2D-Params  & HalfCheetah-Vel   & Meta-World \\
\hline
EWC   & $ 168.43 \pm 649.34 $  &  $  141.77 \pm 29.59 $    & $ -162.55 \pm 97.15 $ & $ 0.42 \pm  0.49$  \\

LwF   & $ 688.67 \pm 306.91 $   &  $  146.13 \pm 14.98$    & $ -82.17 \pm 24.24 $ & $ 0.32  \pm 0.47 $ \\

PackNet   & $ 926.96 \pm  135.71 $  &  $  144.45 \pm 16.64$    & $ -77.06 \pm 27.94 $ & $ 1.0 \pm 0.0 $\\

BC &  $ 992.05 \pm 105.89  $   & $ 161.67 \pm 14.62$  &  $ -31.19 \pm 9.21  $ & $ 1.0 \pm 0.0 $ \\

ClonEx &  $ 1009.90 \pm  67.46 $   & $ 153.26 \pm 13.38 $  &  $ -33.91 \pm  12.89  $ & $ 1.0 \pm 0.0  $ \\

CuGRO & $ \mathbf{ 1036.83 \pm 71.56} $ &  $ \mathbf{173.45 \pm 10.40 } $  &  $ \mathbf{-19.23 \pm  7.99} $ & $ \mathbf{1.0 \pm 0.0 }$ \\ 
\cmidrule[\heavyrulewidth]{1-5}
\end{tabular}
\label{tab:results_all}
\end{table*}

In general, CuGRO consistently achieves performance gain compared to various baselines and different categories of continual learning approaches on CORL.
The superiority is more pronounced as it still outperforms rehearsal-based methods that replay real samples of previous tasks.

\subsection{Ablation Study}\label{exp:ablation}
We compare the performance of CuGRO trained with different replay variants, including: 
1) \textit{Oracle}, it serves as the upper bound with exact replay by assuming a perfect generator. Therefore, Oracle uses real samples of all previous tasks to jointly train the whole model pipeline with new task samples.
2) \textit{Noise}, it considers the opposite case when generated samples do not resemble the real distribution at all.
By omitting the state generator, we generate random noises as replayed state samples that are used to train the behavior generator and multi-head critic.
3) \textit{None}, it naively trains the model pipeline on sequential tasks using new task samples only, without any experience replay.

\begin{figure*}[tb]\centering
 \subfigure[Swimmer-Dir]{\includegraphics[width=0.25\textwidth]{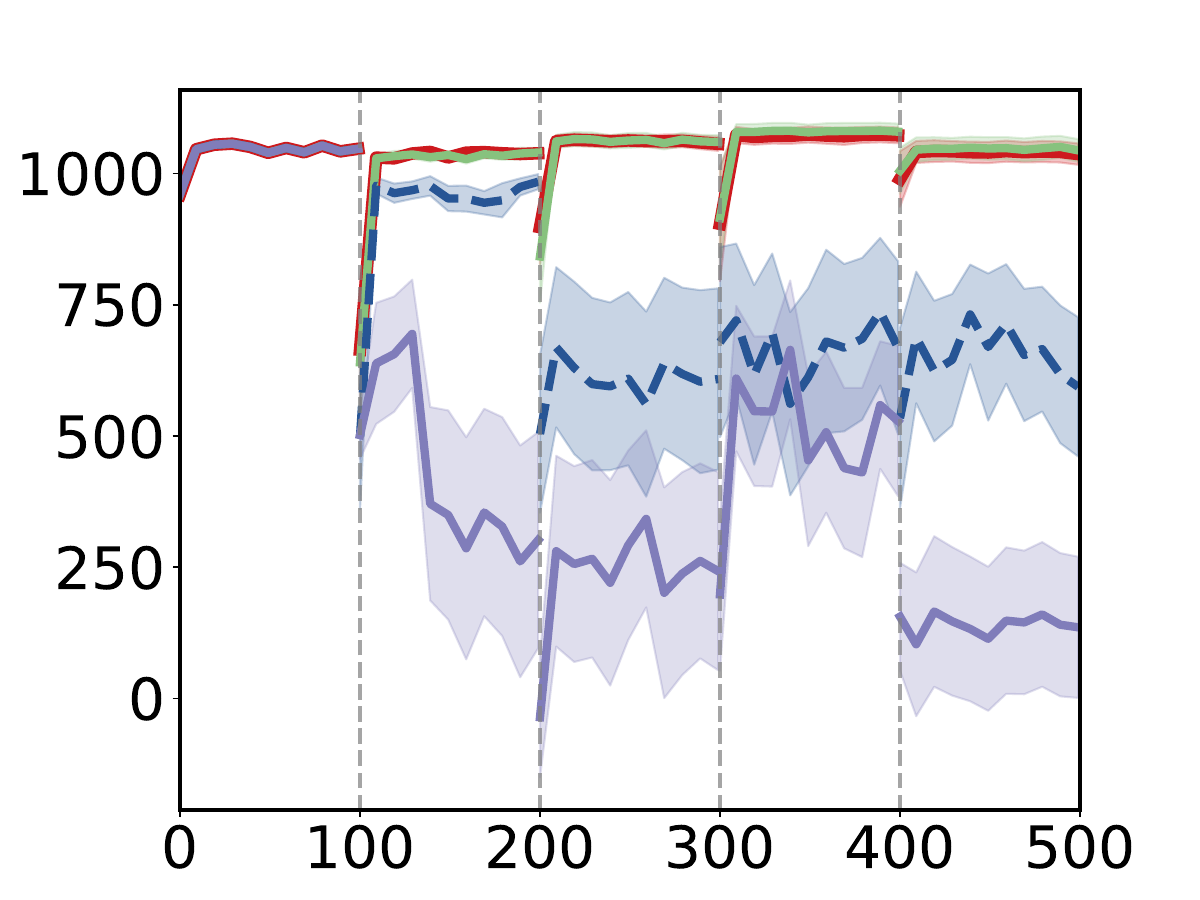}}%
 \subfigure[Walker2D-Params]{\includegraphics[width=0.25\textwidth]{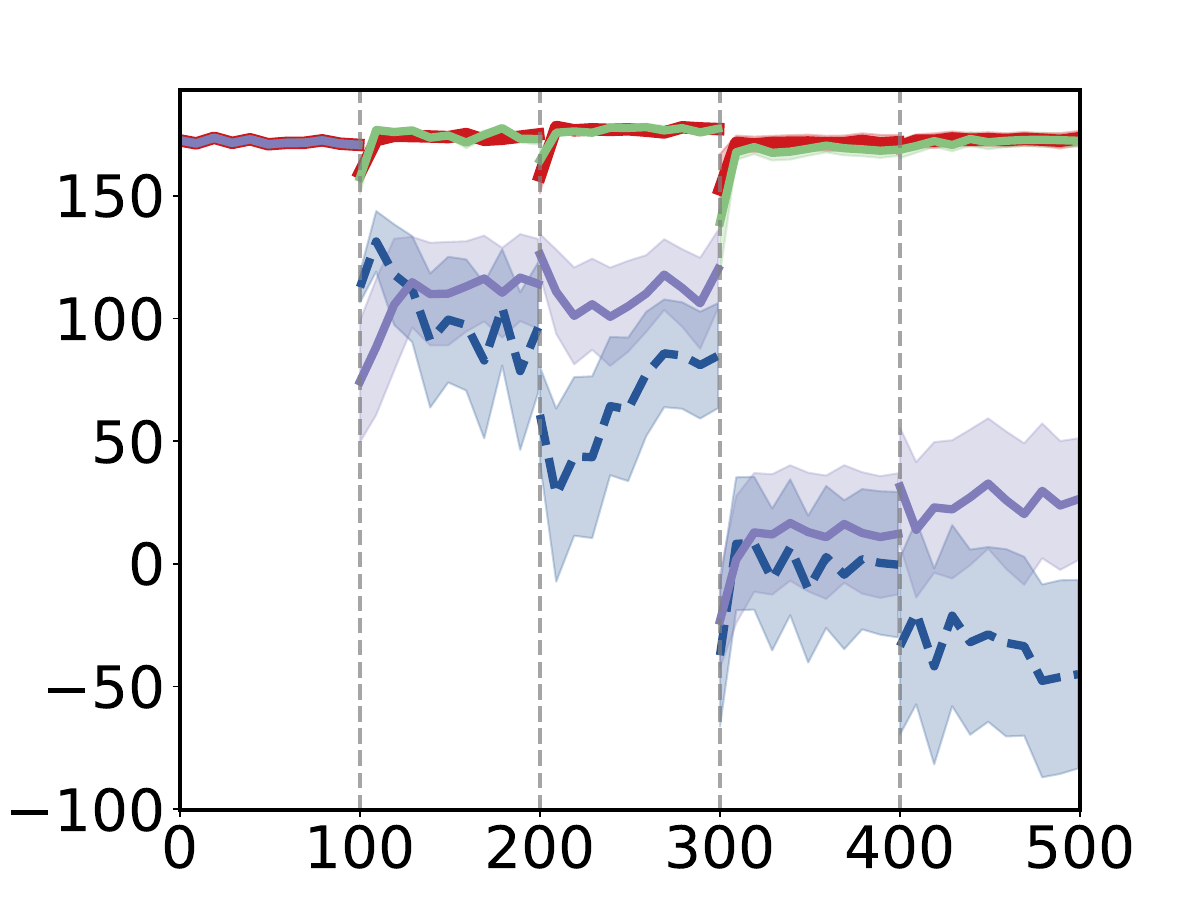}}%
 \subfigure[HalfCheetah-Vel]{\includegraphics[width=0.25\textwidth]{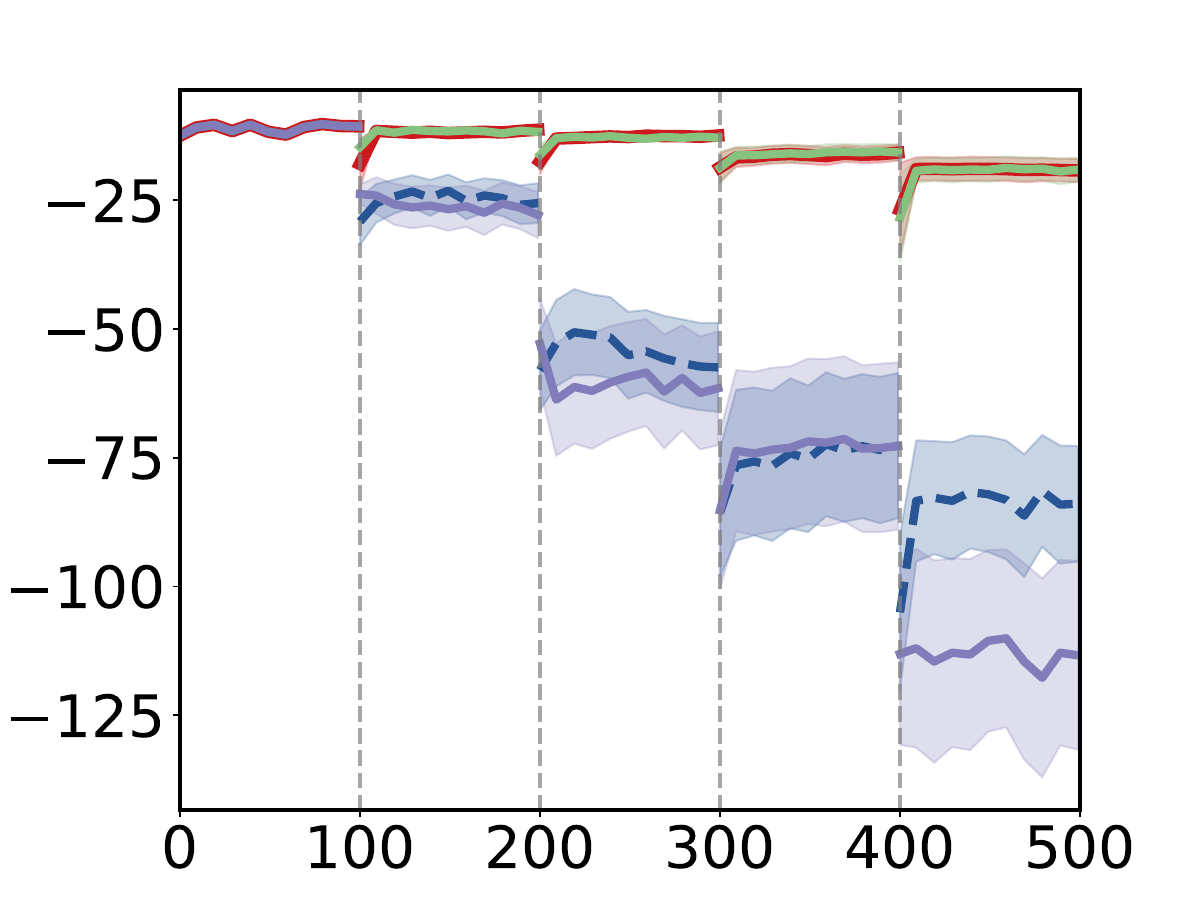}}%
 \subfigure[Meta-World]{\includegraphics[width=0.25\textwidth]{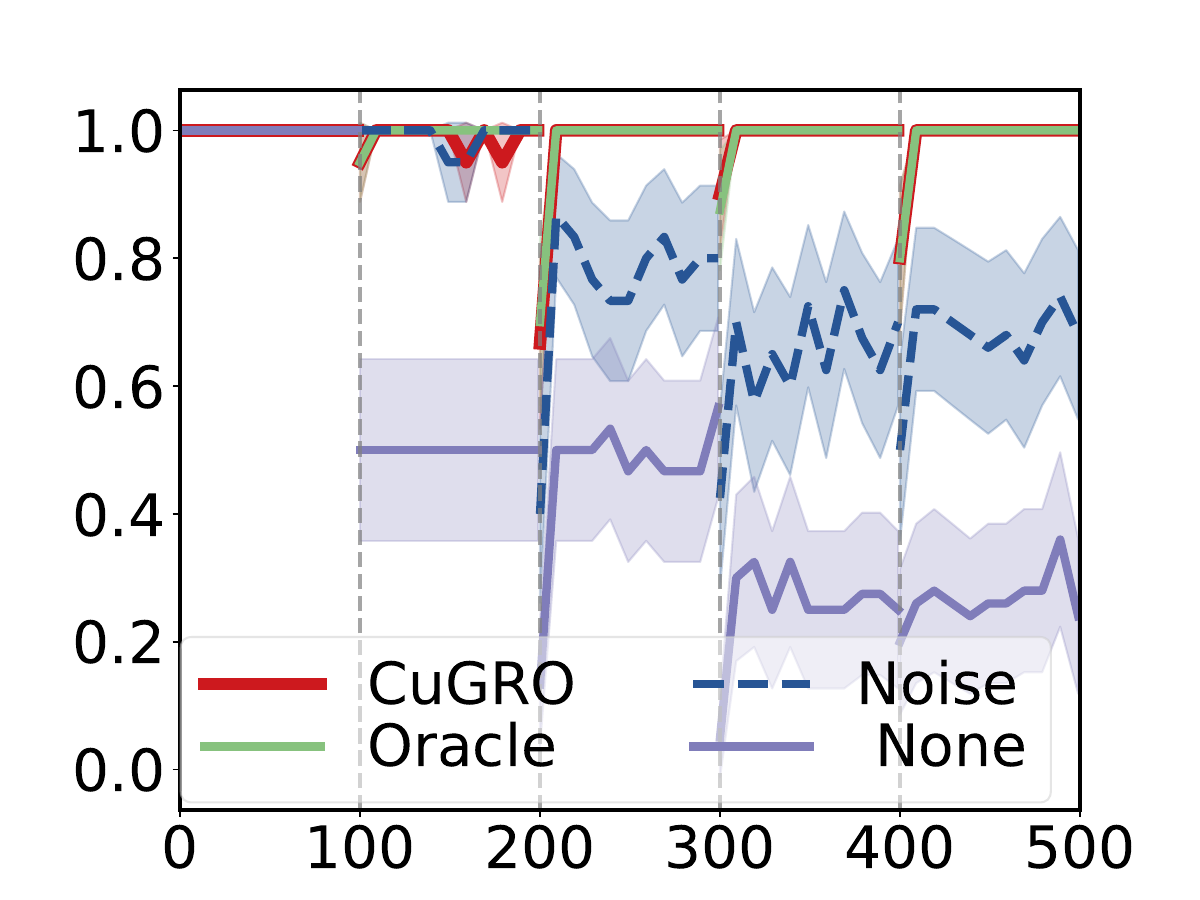}}
 \caption{Performance of CuGRO with different replay variants over cumulative tasks during sequential training.}
 \label{fig:replay_all} 
\end{figure*}

Fig.~\ref{fig:replay_all} shows the sequential training performance of CuGRO with different replay variants, and Table~\ref{tab:replay_all} presents the final performance averaged over all sequential tasks.
A higher return is achieved when the replayed samples better resemble the real data.
Our primary concern is to validate whether CuGRO can realize a high-fidelity replay of the sample space.
Hence, we first look into the performance gap between CuGRO and Oracle to see whether CuGRO can closely approximate the results of using previous ground-truth data.
It can be observed that CuGRO stably preserves the performance of previous tasks, and exhibits almost the same capacity as Oracle.
It demonstrates CuGRO's capability of accurately modeling the mixed data distribution to produce almost full performance compared to real experience replay.
In contrast, both the None and Noise baselines incur severe catastrophic forgetting.
An interesting point is that Noise slightly outperforms None.
It shows that adding random noises to the sample space could augment the feature space with enhanced generalization and robustness, thus helping alleviate forgetting to some extent.

\begin{table*}[!h]
\centering
\setlength{\tabcolsep}{5mm}
\caption{Final performance of CuGRO with different replay variants averaged over all sequential tasks.}
\renewcommand\arraystretch{1}
\begin{tabular}{c|c|c|c|c}
\cmidrule[\heavyrulewidth]{1-5}
Method  & Swimmer-Dir  & Walker2D-Params  & HalfCheetah-Vel   & Meta-World \\
\hline
None    & $ 135.61 \pm 473.01 $  &  $ 26.27 \pm  87.44 $    & $ -113.38 \pm 64.54  $  & $  0.24 \pm 0.43  $ \\
 
Noise &  $ 593.48 \pm 466.78 $    & $ -45.05 \pm 135.20 $  &  $ -83.94 \pm 39.51 $ & $  0.68 \pm 0.47$\\

Oracle &  $ \mathbf{1043.87 \pm 75.69} $     & $  172.45 \pm 8.82 $  &  $ \mathbf{-19.05 \pm 8.14} $ & $  \bm{1.0 \pm 0.0} $ \\

CuGRO    &  $ 1036.83 \pm 71.56 $   &  $ \mathbf{173.45 \pm 10.40} $  &  $ -19.23 \pm 7.99$ & $ \bm{1.0 \pm 0.0} $ \\ 

\cmidrule[\heavyrulewidth]{1-5}
\end{tabular}
\label{tab:replay_all}
\end{table*}

\subsection{Superiority of Diffusion-based Replay}\label{exp:difusion}
\begin{figure}[!h]\centering
 \subfigure[HalfCheetah-Vel]{\includegraphics[width=0.249\textwidth]{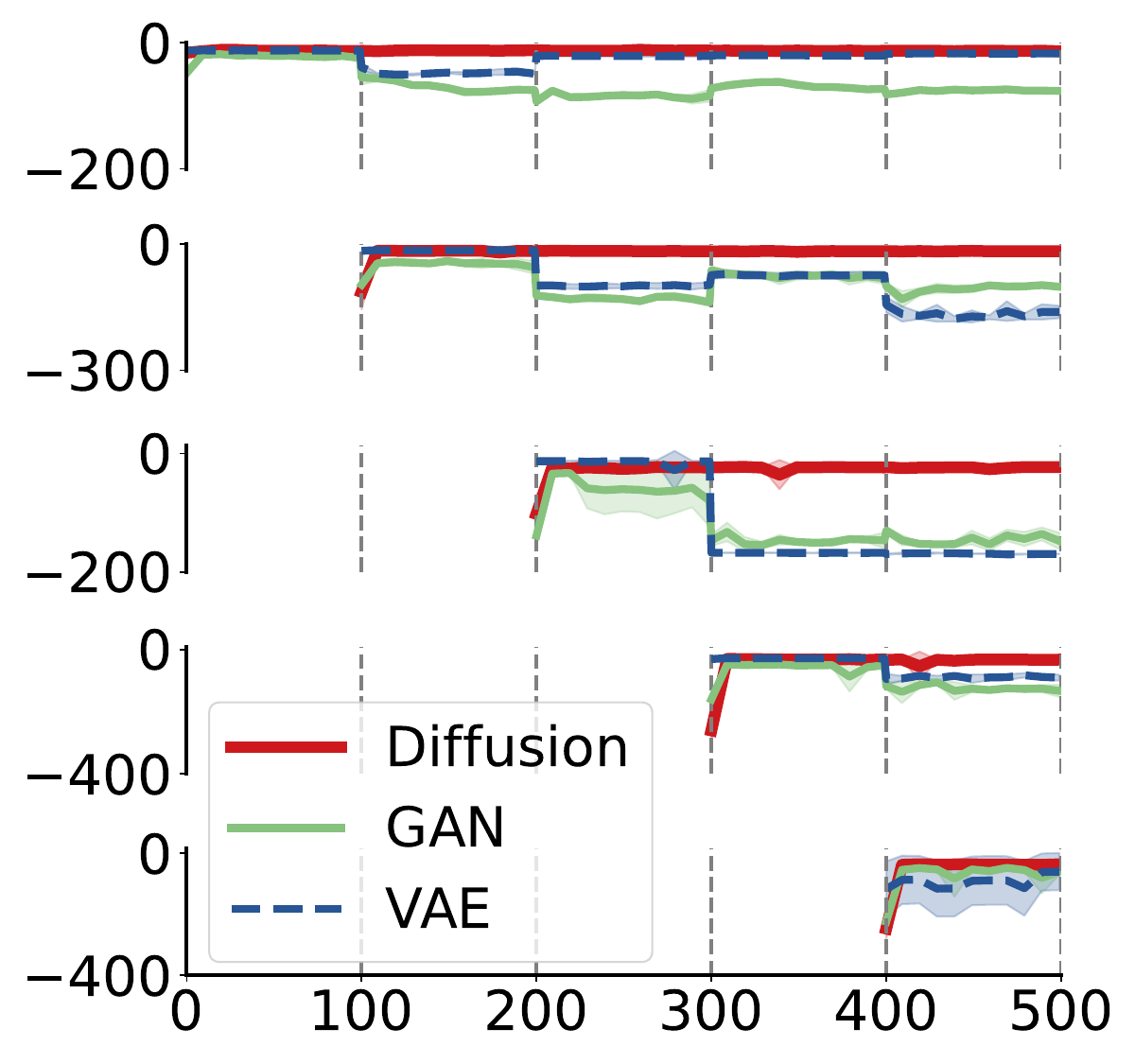}}%
 \subfigure[Meta-World]{\includegraphics[width=0.249\textwidth]{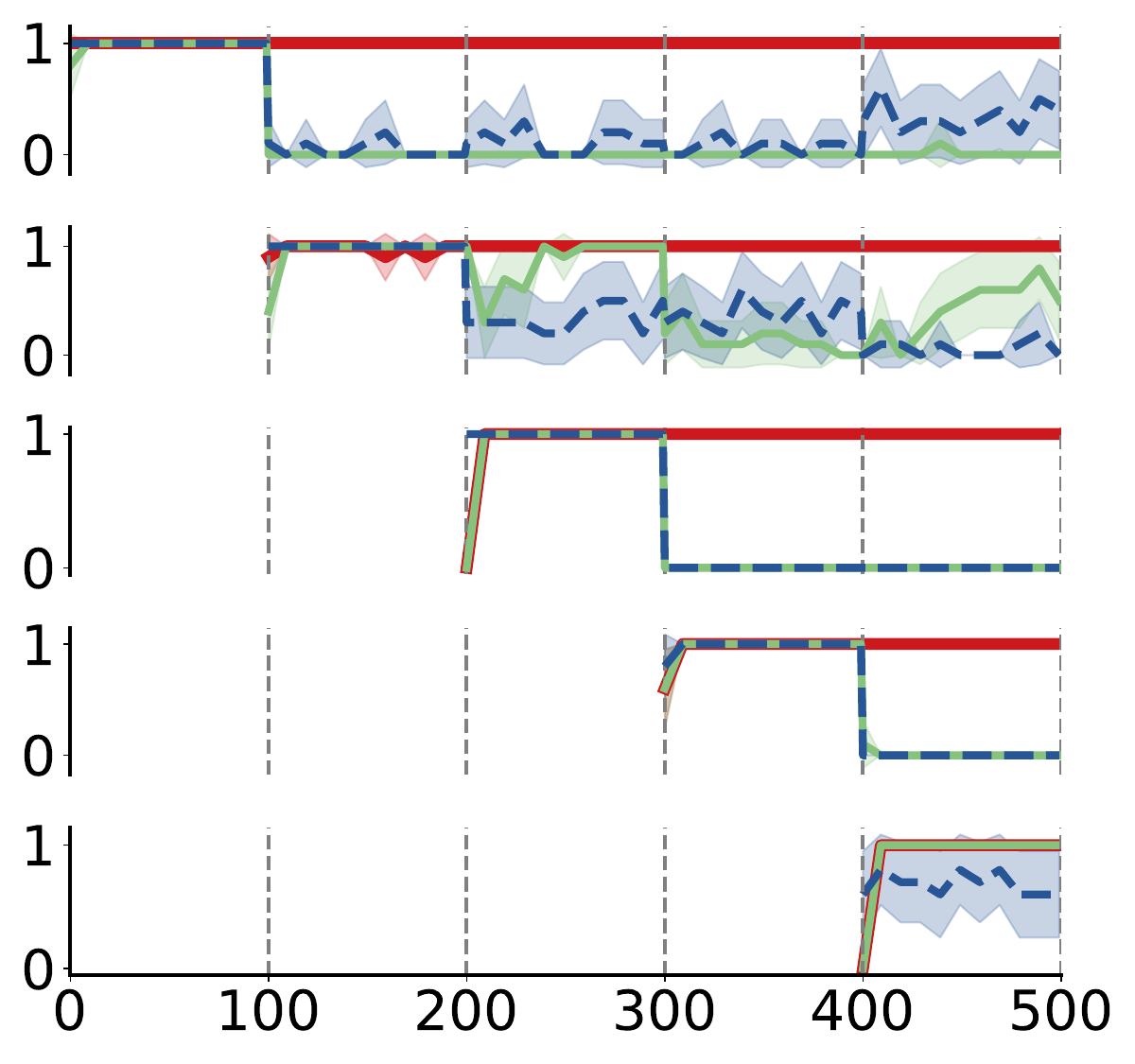}}%
 \caption{Performance of CuGRO with different generative models over cumulative tasks during sequential training.}
 \label{fig:model_part} 
\end{figure}

For generative replay, we leverage advances in diffusion models to mimic the state and behavior distributions of previous tasks.
We are also interested in observing the performance when incorporating other generative models like VAEs or GANs.
Hence, we replace the diffusion-based generators with conditional VAEs~\citep{sohn2015learning} and conditional GANs~\citep{mirza2014conditional} for conditioned state and behavior generation.
Fig.~\ref{fig:model_part} presents the sequential training performance of CuGRO with different generators, and Table~\ref{tab:model_part} shows the corresponding final performance averaged over all sequential tasks.
More results can be found in Appendix~E.
When using VAEs or GANs as generators, the performance of previous tasks degrades with a significant degree of forgetting.
The GANs variant obtains slightly better performance compared to VAEs.
It aligns with the advancement in the generative modeling community, where it is generally observed that GANs exhibit greater generation capacity than VAEs.
In summary, the results validate the superiority of CuGRO compared to using other generative models, and also verify our research motivation of leveraging powerful diffusion models to model states and corresponding behaviors with high fidelity.

\begin{table}[!h]
\centering
\setlength{\tabcolsep}{4mm}
\caption{Final performance of CuGRO with different generative models averaged over all sequential tasks.}
\renewcommand\arraystretch{1}
\begin{tabular}{c|c|c}
\cmidrule[\heavyrulewidth]{1-3}
Method  &  HalfCheetah-Vel   & Meta-World \\
\hline
VAE  &  $ -100.04 \pm 70.69 $ & $ 0.20 \pm 0.40 $  \\  
GAN   & $ -104.02 \pm 36.99 $  & $ 0.34 \pm 0.47 $ \\ 
Diffusion   &  $ \mathbf{-19.23 \pm 7.99} $ & $\mathbf{ 1.00 \pm 0.00} $ \\ 
\cmidrule[\heavyrulewidth]{1-3}
\end{tabular}
\label{tab:model_part}
\end{table}

\begin{figure}[tb]\centering
 \subfigure[HalfCheetah-Vel]{\includegraphics[width=0.249\textwidth]{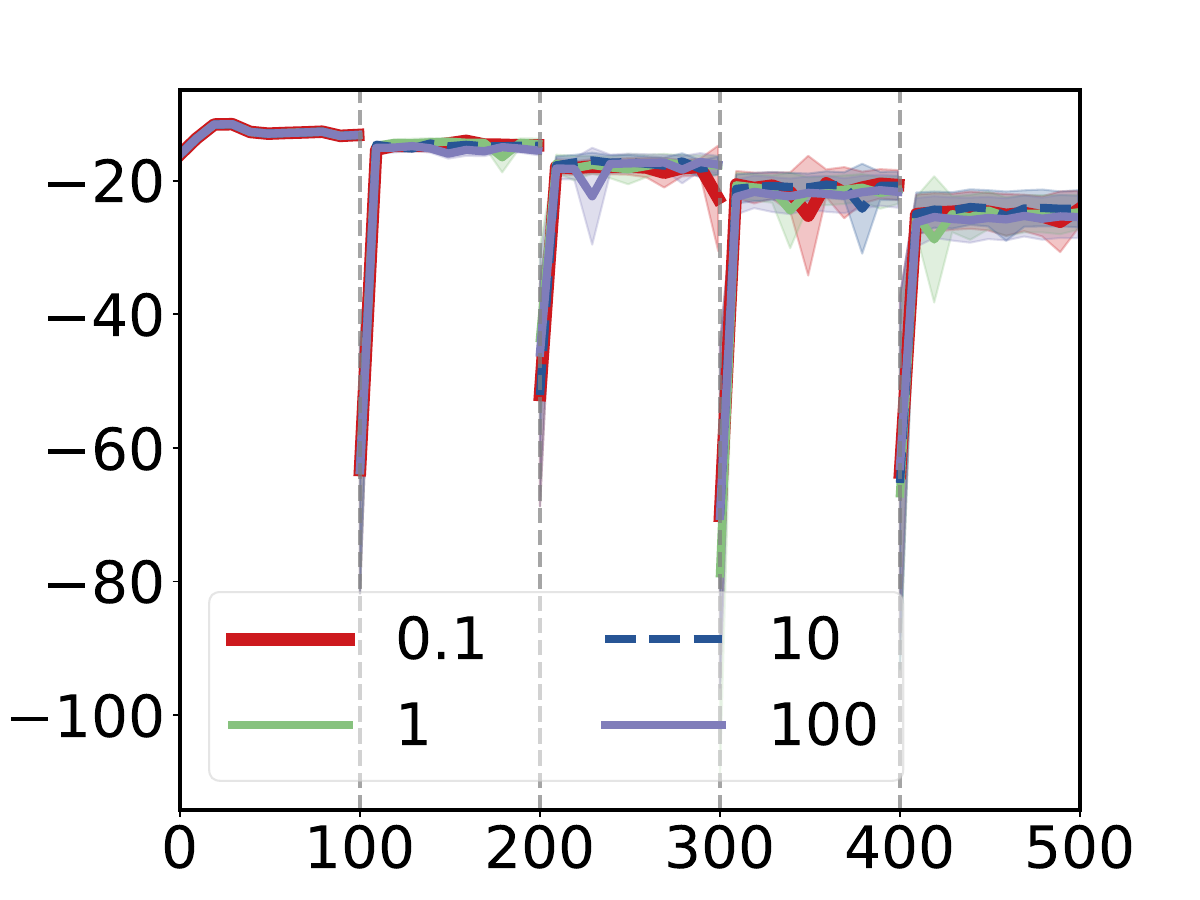}}%
 \subfigure[Meta-World]{\includegraphics[width=0.249\textwidth]{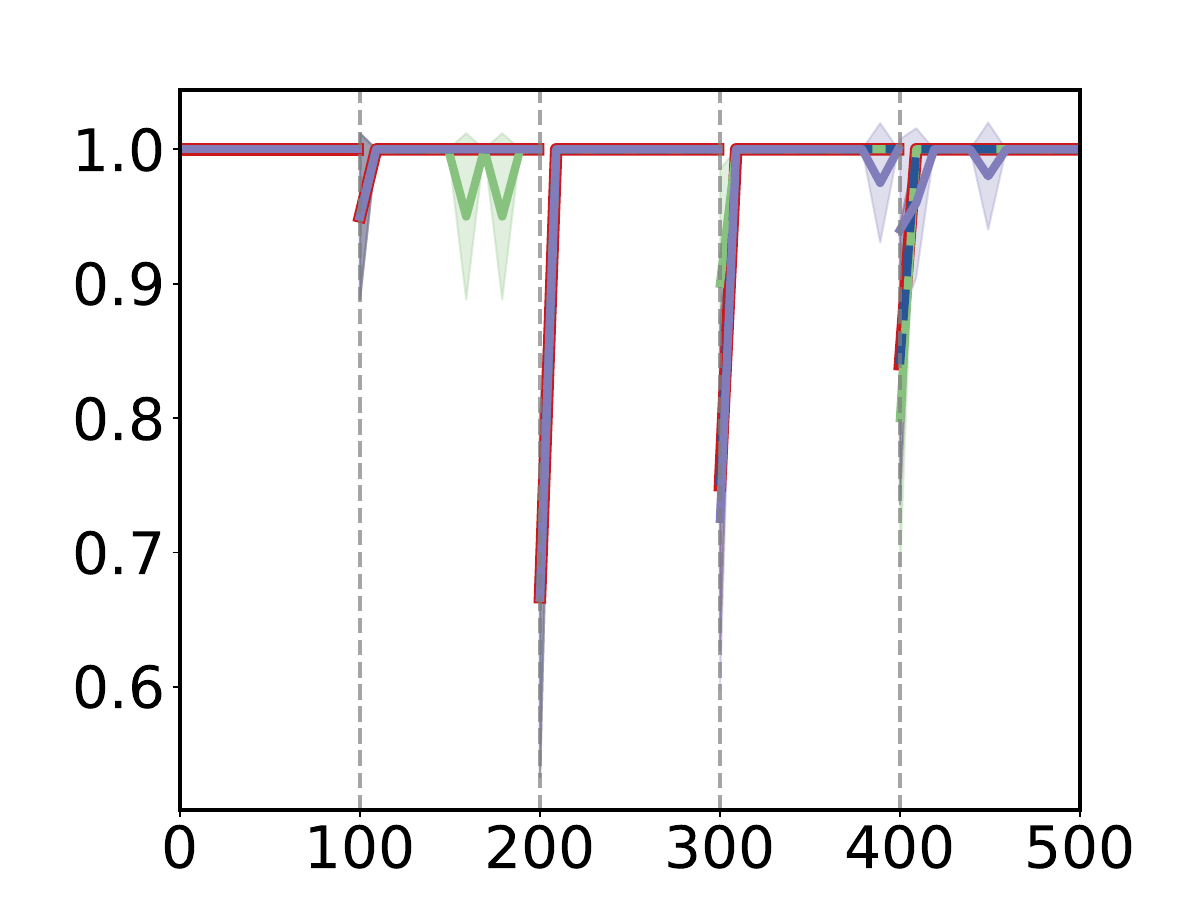}}%
 \caption{Performance of CuGRO with varying coefficients $\lambda$ over cumulative tasks during sequential training.}
 \label{fig:hyper_part} 
\end{figure}

\begin{table}[tb]
\centering
\setlength{\tabcolsep}{5mm}
\caption{Final performance of CuGRO with varying coefficients $\lambda$ averaged over all sequential tasks.}
\renewcommand\arraystretch{1}
\begin{tabular}{c|c|c}
\cmidrule[\heavyrulewidth]{1-3}
$\lambda$  & HalfCheetah-Vel   & Meta-World \\
\hline
0.1   &  $ -19.80 \pm 8.53 $  &  $ 1.00 \pm 0.00 $  \\ 
1   &   $  \mathbf{ -19.23 \pm 7.99} $ & $ 1.00 \pm 0.00 $ \\ 
10 &   $ -19.37 \pm 8.28 $ & $ 1.00 \pm 0.00 $  \\ 
100 &   $ -19.66 \pm 8.58 $  &  $ 1.00 \pm 0.00 $ \\ 
\cmidrule[\heavyrulewidth]{1-3}
\end{tabular}
\label{tab:hyper_part}
\end{table}

\subsection{Hyperparameter Analysis}
The coefficient $\lambda$ in Eq.~(\ref{train_critic}) is a key factor that balances the learning of the expanded new head (plasticity) and the behavior cloning of previous knowledge (stability).
With $\lambda=1$, CuGRO is performant in all evaluation domains.
Moreover, we conduct experiments to analyze the influence of coefficient $\lambda$ on CuGRO's performance.
Fig.~\ref{fig:hyper_part} and Table~\ref{tab:hyper_part} show the sequential training performance and average final performance with varying coefficients, and more results can be found in Appendix~F.
A higher coefficient corresponds to more emphasis on replaying previous tasks.
Results show that CuGRO achieves good stability and robustness as the performance is insensitive to the pre-defined balancing coefficient.
It again verifies the effectiveness of the multi-head critic for tackling task diversity and the efficiency of behavior cloning for mitigating forgetting.

\section{Related work}
\textbf{Offline RL} imposes new challenges including overestimation of out-of-distribution (OOD) actions and accumulated extrapolation errors~\citep{levine2020offline}.
Policy regularization tackles the problem by constraining policy distribution discrepancy to avoid visiting state-action pairs less covered by the dataset~\citep{kumar2019stabilizing,fujimoto2021minimalist,ran2023policy}. 
Pessimistic value-based approaches learn a conservative Q-function for unseen actions to discourage selecting OOD actions~\citep{kostrikov2021offline1, kostrikov2021offline2,yang2022rorl}.
An alternative is to learn a constrained actor with advantage-weighted regression to implicitly enforce a constraint on distribution shift and overly conservative updates~\citep{peng2019advantage,nair2020awac, wang2020critic}.
Recently, some studies~\citep{kumar2019stabilizing,zhou2021plas,chen2022lapo} attempt to leverage advances in generative models, such as VAEs~\citep{fujimoto2019off} or diffusion models~\citep{lu2023contrastive}, to model offline datasets for model-based planning~\citep{janner2022planning} or to directly model the policy with regularization~\citep{wang2023diffusion}.

\textbf{Continual RL} aims to facilitate forward transfer and mitigate forgetting of previous tasks.
Regularization-based approaches impose additional terms on learning objectives to penalize large updates on weights that are important for previous tasks, such as  EWC~\citep{kirkpatrick2017overcoming}, OWM~\citep{zeng2019continual}, and PC ~\citep{kaplanis2019policy}.
With limited resources, comprising regularization might result in a proficiency compromise between previous and new tasks.
Parameter isolation approaches learn to mask a sub-network for each task in a shared network~\citep{kang2022forget,konishi2023parameter}, or dynamically expand model capacity to incorporate new information~\citep{kessler2022same,wang2022dirichlet}, to prevent forgetting by keeping parameters of previous tasks unaltered.
However, they could potentially suffer from over-consumption of network capacity as each task monopolizes and consumes some amount of resources.

Using the idea of episodic memory, rehearsal-based methods~\citep{jeeveswaran2023birt} store samples from previous tasks that are reproduced for interleaving online updates when learning a new task~\citep{isele2018selective,rolnick2019experience,wolczyk2022disentangling}.
~\citep{gai2023oer} proposes a model-based experience selection scheme to alleviate the distribution shift between the stored memory and samples from the learned policy.
~\citep{huang2024solving} employs distillation and selective rehearsal to enhance continual learning via multi-head decision transformers.
Nonetheless, they might not be viable in real-world scenarios due to requiring a large working memory and involving several ramifications like data privacy or security concerns. 
To bypass the memory burden, generative replay methods~\citep{shin2017continual} turn to use generative models to mimic older parts of the data distribution and sample synthetic data for rehearsal~\citep{qi2023better}.
Classical approaches adopt VAEs~\citep{ketz2019continual} or GANs~\citep{zhai2019lifelong} as the generator, and recent work suggests that powerful diffusion models can be employed with high-quality image synthesis~\citep{gao2023ddgr}. 
~\citep{ketz2019continual} continually trains a world model by interleaving generated episodes of past experiences.
~\citep{daniels2022model} uses a VAE generator to replay observation-action pairs in Starcraft II and Minigrid domains.

\textbf{Diffusion models}~\citep{sohl2015deep} achieve great success in synthesizing high-fidelity images~\citep{dhariwal2021diffusion}, audios~\citep{kim2022guided}, and videos~\citep{ho2022imagen}. 
The unconditional diffusion model does not require additional signals and learns the reverse process by estimating the noise at each step, such as DDPM~\citep{ho2020denoising,nichol2021improved}.
To accelerate sampling, DDIM~\citep{song2020denoising} generalizes the Markovian forward process of DDPM to a class of non-Markovian ones that lead to the same training objective.
The conditional diffusion model relies on various source signals, e.g., class labels in image tasks, to generate corresponding data.
~\citep{dhariwal2021diffusion} conditions on the gradient from a pretrained classifier to trade off diversity for producing high-fidelity samples.
Further, ~\citep{ho2022classifier} proposes classifier-free guidance that jointly trains a conditional and an unconditional diffusion model to attain a trade-off between sample quality and diversity.
In this paper, we build our method on top of the diffusion model in~\citep{song2021score} that enables new sampling methods and further extends the capabilities of score-based generative models through the lens of stochastic differential equations.

\section{Conclusion and Discussion}
In the paper, we tackle the CORL challenge via a dual generative replay framework that leverages advances in diffusion models to model states and behaviors with high fidelity, allowing the continual learning policy to inherit the distributional expressivity.
The dual state and behavior generators are continually trained to model a progressive range of diverse behaviors via mimicking a mixed data distribution of real samples and replayed ones from previous generators.
Experiments on classical benchmarks verify the superiority of our method and its high-fidelity replay of the sample space.
Though, our method requires two diffusion models to synthesize replayed samples, which could be further improved by sampling acceleration methods or developing one diffusion model for unifying the state and behavior modeling.
We leave these directions as future work.

\clearpage
\bibliography{example_paper}
\bibliographystyle{IEEEtran}
\clearpage
\appendix
\onecolumn

\section*{Appendix A. Algorithm Summary}\label{algo_sum}
Based on the implementations in Section 3, we summarize the brief procedure of our method in Algorithm~\ref{algo}. 

\begin{algorithm}[h]
\caption{Continual offline RL via diffusion-based dual generative replay}
\label{algo}
\begin{algorithmic}
\STATE {\bfseries Inputs:} $(M_1,..., M_K,...)$: sequential tasks;~~~~~ $\mathcal{D}_{\mu_k}$: offline dataset of task $M_k$, $k=1,...,K$; \\
~~~~~~~~~~~~ $\mu_{\bm{\phi}}(\bm{a}|\bm{s})$: state-conditioned behavior generative model with parameters $\bm{\phi}$; \\
~~~~~~~~~~~~ $p_{\bm{\varphi}}(\bm{s}|k)$: task-conditioned state generative model with parameters $\bm{\varphi}$; \\
~~~~~~~~~~~~ $Q_{\bm{\theta}}^k(\bm{s},\bm{a})$: multi-head critic with parameters $\bm{\theta}$.

\FOR{Task $K=1,2,...$}
    \IF{$K=1$}
        \STATE Train the nominal behavior generative model $\bm{\phi}_1$ with $\mathcal{D}_{\mu_1}$ using Eq.~(\ref{action_update}) 
        \STATE Train the nominal action evaluation model $\bm{\theta}_1$ with $\mathcal{D}_{\mu_1}$ using Eq.~(\ref{qlearning}) 
        \STATE Train the nominal state generative model $\bm{\varphi}_1$ with $\mathcal{D}_{\mu_1}$ using Eq.~(\ref{noise_state})
    \ELSE
        \STATE Initialize dataset: $\mathcal{D}=\mathcal{D}_{\mu_K}$
        \FOR{$k = 1$ to $K-1$} 
            \STATE Generate state samples: $\hat{\bm{s}}_k\sim p_{\bm{\varphi}_{K-1}}(\bm{s}|k)$ 
            \STATE Generate corresponding action samples: $\hat{\bm{a}}_k\sim \mu_{\bm{\phi}_{K-1}}(\bm{a}|\hat{\bm{s}}_k)$   
            \STATE Construct pseudo dataset: $\hat{\mathcal{D}}_k = \sum (\hat{\bm{s}}_k, \hat{\bm{a}}_k)$ 
            \STATE Annotate the Q-function of pseudo state-action pairs as $Q_{\bm{\theta}_{K-1}}^k(\hat{\bm{s}}_k, \hat{\bm{a}}_k)$
            \STATE Interleave pseudo samples with real ones: $\mathcal{D} = \mathcal{D} \cup \hat{\mathcal{D}}_k$
        \ENDFOR
        \STATE Initialization of models: $(\bm{\varphi}_K, \bm{\phi}_K, \bm{\theta}_K)\leftarrow (\bm{\varphi}_{K-1}, \bm{\phi}_{K-1}, \bm{\theta}_{K-1})$ 
        \STATE Update state generative model $\bm{\varphi}_K$ with $\mathcal{D}$ using Eq.~(\ref{train_state}) 
        \STATE Update behavior generative model $\bm{\phi}_K$ with $\mathcal{D}$ using Eq.~(\ref{train_behavior}) 
        \STATE Update the multi-head critic with $\mathcal{D}$ and $\{Q_{\bm{\theta}_{K-1}}^k(\hat{\bm{s}}_k, \hat{\bm{a}}_k)\}_{k=1}^{K-1}$ using Eq.~(\ref{train_critic}) 
    \ENDIF
\ENDFOR

\end{algorithmic}
\end{algorithm}

\section*{Appendix B. Details of Evaluation Benchmarks and Data Collection}
This section introduces the studied MuJoCo and Meta-World benchmarks in detail, as well as the offline dataset collection process for all evaluation domains.

\textbf{Evaluation Benchmarks.}
The multi-task MuJoCo control testbed is a classical benchmark commonly used in continual RL, multi-task RL, and meta-RL. 
This testbed concludes two environments with reward function changes and one environment with transition dynamics changes as
\begin{itemize}
    \item \textit{Swimmer-Dir}: A swimmer robot needs to move in a given direction.
    Tasks differ in reward functions, which describe the target direction.
    We randomly sample five target directions in 2D space to form the sequential tasks.
    
    \item \textit{Walker2D-Params}: A planar walker robot needs to move forward as fast as possible.
    Tasks differ in transition dynamics. For each task, the physical parameters of body mass, inertia, damping, and friction are randomized. The agent should adapt to the varying environment dynamics to accomplish the task. 
    We randomly sample five sets of physical parameters to construct the continual learning task.
    
    \item \textit{HalfCheetah-Vel}: A planar cheetah needs to run forward at a goal velocity.
    Tasks differ in reward functions, which describe the target velocity of the agent.
    We randomly sample five target velocities in the range of $[0,2]$ to form the sequential tasks.
\end{itemize}

\begin{figure*}[!h]\centering
 \subfigure[coffee-button]{\includegraphics[width=0.2\textwidth]{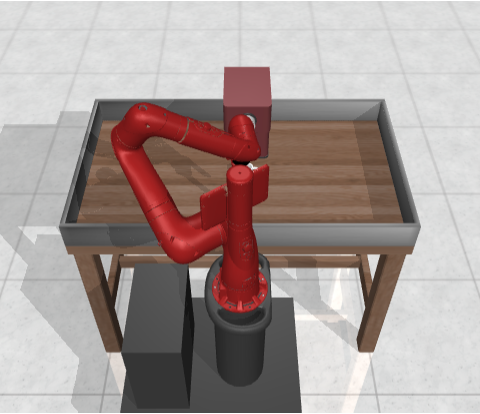}}%
 \subfigure[drawer-close]{\includegraphics[width=0.2\textwidth]{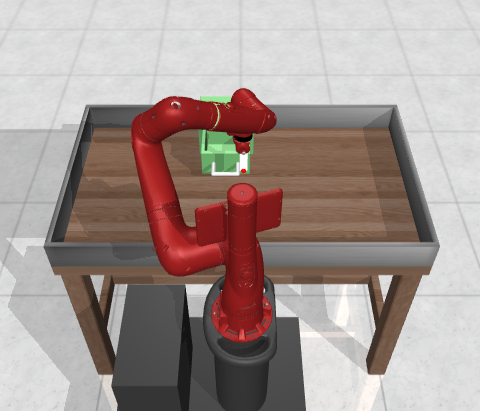}}%
 \subfigure[plate-slide-back]{\includegraphics[width=0.2\textwidth]{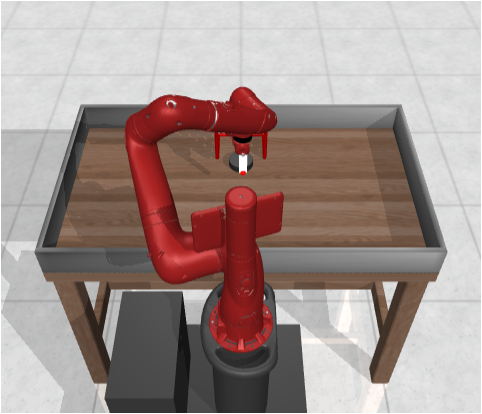}}%
 \subfigure[window-close]{\includegraphics[width=0.2\textwidth]{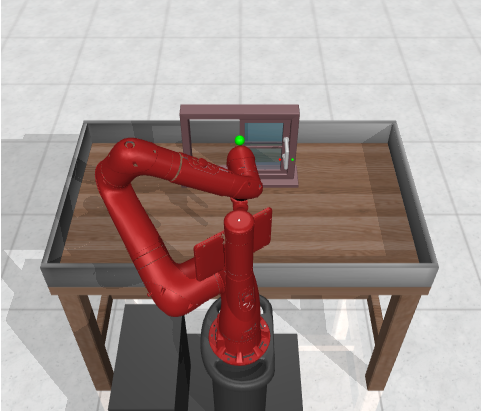}}%
\subfigure[window-open]{\includegraphics[width=0.2\textwidth]{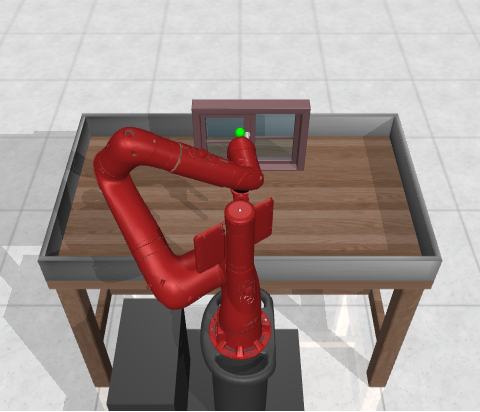}}%
\caption{Visualization of the selected five tasks in the Meta-World testbed.}
\label{fig:metaworld} 
\end{figure*}

The Meta-World benchmark~\citep{yu2020meta} contains a suite of $50$ diverse simulated manipulation tasks with everyday objects, all of which are contained in a shared, table-top environment with a simulated Sawyer arm. 
By providing a large set of distinct tasks that share a common environment and
control structure, Meta-World allows researchers to test the learning capabilities of the continual RL, multi-task RL, and meta-RL methods, and help to identify new research avenues to improve the current approaches.
From this testbed, we sample five different tasks as shown in Fig.~\ref{fig:metaworld} and arrange them in a fixed order to construct a continual learning sequence as
\begin{itemize}
    \item \textit{coffee-button}: Push a button on the coffee machine. Randomize the position of the coffee machine.

    \item \textit{drawer-close}: Push and close a drawer. Randomize the drawer positions.

    \item \textit{plate-slide-back}: Slide a plate into a cabinet. Randomize the plate and cabinet positions.

    \item \textit{window-close}: Push and close a window. Randomize window positions.

    \item \textit{window-open}: Push and open a window. Randomize window positions.
\end{itemize}

\textbf{Data Collection.}
For each evaluation domain, we choose five tasks to execute in a sequence and train a single-task policy independently for each task.
We use soft actor-critic (SAC)~\citep{haarnoja2018soft} for the Swimmer-Vel, Walker2D-Params, and Meta-World domains.
We use the TD3 algorithm~\citep{fujimoto2018addressing} for the HalfCheetah-Vel domain, since it is observed that TD3 provides a more stable learning process across various HalfCheetah-Vel tasks~\citep{mitchell2021offline}.
For each task in each domain, we set the time horizon in a learning episode as $200$ and the number of training steps as $1M$.
We collect three benchmarks for each task: 
1) Replay: Save the complete replay buffer from the entire lifetime of training to construct the dataset for the downstream offline RL algorithms.
2) Medium: Construct the dataset with a size of 100000 from medium-quality policy.
3) Expert: Construct the dataset with a size of 100000 from expert-quality policy.
Table~\ref{tab1} and Table~\ref{tab2} list the main hyper-parameters for the SAC and TD3 algorithms during offline data collection in all evaluation domains, respectively.

\begin{table*}[!h]
\setlength{\tabcolsep}{5mm}
\renewcommand\arraystretch{1.2}
    \centering
    \caption{Hyperparameters for SAC in the data collection phase.}
    \vspace{0.5em}
    \begin{tabular}{cccc}
        \toprule
    Parameter  & Swimmer-Vel & Walker2D-Params &  Meta-World  \\
        \midrule	
    Optimizer            & Adam & Adam & Adam \\ 
    Value learning rate  & $1e-4$  & $1e-4$ & $1e-3$ \   \\
    Policy learning rate & $1e-4$ & $1e-4$ & $1e-3$ \\\
    Automating temperature parameter    & $0.2$ & $0.2$ & $0.2$ \\
    Temperature parameter learning rate & $1e-4$ & $1e-4$ & $1e-3$ \ \\
    Batch size           & $256$   & $256$ & $128$ \\
    Neurons per hidden layer & $256$  & $256$  & $256$ \\
    Number of hidden layers  & $1$  & $1$  & $3$\\	
    Discount factor        & $0.99$ & $0.99$   & $0.99$ \\
    Target network update rate        &  $0.01$  &  $0.01$ &  $0.005$ \\
    \bottomrule
    \end{tabular}
    \label{tab1}
\end{table*}

\begin{table*}[!h]
\setlength{\tabcolsep}{5mm}
\renewcommand\arraystretch{1.2}
    \centering
    \caption{Hyperparameters for TD3 in the data collection phase.}
    \vspace{0.5em}
    \begin{tabular}{cc}
        \toprule
    Parameter  & HalfCheetah-Vel  \\
        \midrule	
    Optimizer            & Adam  \\ 
    Value learning rate  & $1e-3$ \\
    Policy learning rate & $1e-4$ \\
    Batch size           & $256$  \\
    Neurons per hidden layer & $256$  \\
    Number of hidden layers  & $1$ \\	
    Discount factor        & $0.99$  \\
    Target network update rate        &  $0.01$ \\
    Range to clip target policy noise       &   $0.5$ \\
    Policy noise        &   $0.2$   \\
    Exploration noise        &   $0.1$  \\
    Frequency of delayed policy updates        &   $2$ \\
    \bottomrule
    \end{tabular}
    \label{tab2}
\end{table*}

 \section*{Appendix C. Details of Baseline Methods}\label{baselines}
This section gives the details of the representative five baselines, including three rehearsal-based approaches, one regularization-based approach, and one parameter isolation-based approach. 
These baselines are thoughtfully selected to encompass the three primary categories of continual learning methods.
Also, since our method CuGRO belongs to the rehearsal-based category, we adopt more approaches from this kind as our baselines.
The baselines are introduced as follows.

\begin{itemize}
    \item \textbf{ClonEx}~\citep{wolczyk2022disentangling}, is a rehearsal-based continual learning method that incorporates a behavioral constraints term into the loss function of the policy network.
The network employs a multi-head architecture, wherein each task has a dedicated output header. It features a shared backbone with 4 hidden layers and task-specific output layers serving as headers. 
At the end of each task, 10,000 state samples are randomly selected from the dataset and labeled using the outputs (distribution) of the current (trained) network as ``expert" data. 
In the subsequent tasks, auxiliary losses based on the KL divergence are added to the policy objective to imitate this expert data.  Hyperparameter selection aligns with the optimal values found in the original paper of ClonEx.

\item \textbf{Behavior Cloning (BC)}~\citep{isele2018selective}, is a rehearsal-based continual learning method that incorporates a behavioral cloning term into the policy network's loss function. Similar to ClonEx, the actor utilizes a multi-head network architecture. 
Upon the completion of each task, 10,000 state samples are randomly selected from the dataset and labeled with the optimal actions determined by the current (trained) network, forming `expert' data. 
For subsequent tasks, auxiliary losses based on the L2 norm are introduced to the policy objective to emulate this expert data.

\item \textbf{LwF}~\citep{li2017learning}, is a rehearsal-based continual learning method that trains the network on new task data while preserving performance on previous tasks. 
Like the previously mentioned methods, it employs a multi-head network architecture. 
However, instead of storing real samples from previous tasks, LwF randomly selects state samples from the current dataset of the new task. 
For the same state input, the outputs of the old tasks' heads in the new actor (during training) are constrained to remain close to those of the corresponding heads in the old actor (trained). 
In subsequent tasks, auxiliary losses based on KL divergence are added to the policy objective to approximate the behavior of the old actor.

\item \textbf{PackNet}~\citep{mallya2018packnet}, is a continual learning method that utilizes parameter isolation to accommodate multiple tasks within a single network. 
This is achieved by iterative pruning and retraining the network. The actor in PackNet is a single-head network, PackNet manages the network's parameters through an iterative process of pruning and retraining, thereby `making space' for new tasks. After the completion of each task, a portion of the network's parameters is frozen to prevent performance degradation on previous tasks.

\item \textbf{EWC}~\citep{kirkpatrick2017overcoming}, is a classical regularization-based continual learning method that constrains changes to critical parameters through the Fisher information matrix. 
By assigning higher `importance' to parameters crucial for previous tasks, EWC allows the network to retain the memory of old tasks as it learns new ones. 
This is achieved by adding an additional regularization term to the loss function, which penalizes significant changes to these important parameters.
\end{itemize}

As the CORL challenge is an understudied problem, existing continual RL methods are usually explored within online settings.
Conventional online RL algorithms generally suffer from significant performance degradation when simply deployed to offline settings.
The challenges include the overestimation of out-of-distribution actions and accumulated extrapolation errors due to the distribution discrepancy between the behavior and target policies~\citep{levine2020offline}.
For a fair comparison, we modify the five baselines to offline settings and implement them based on the classical offline RL algorithm AWAC~\citep{nair2020awac}.

\section*{Appendix D. Inplementation Details of CuGRO}\label{cugro}
\begin{figure}[!h]\centering
\subfigure[The network architecture of the behavior generative model.]{\includegraphics[width=0.9\textwidth]{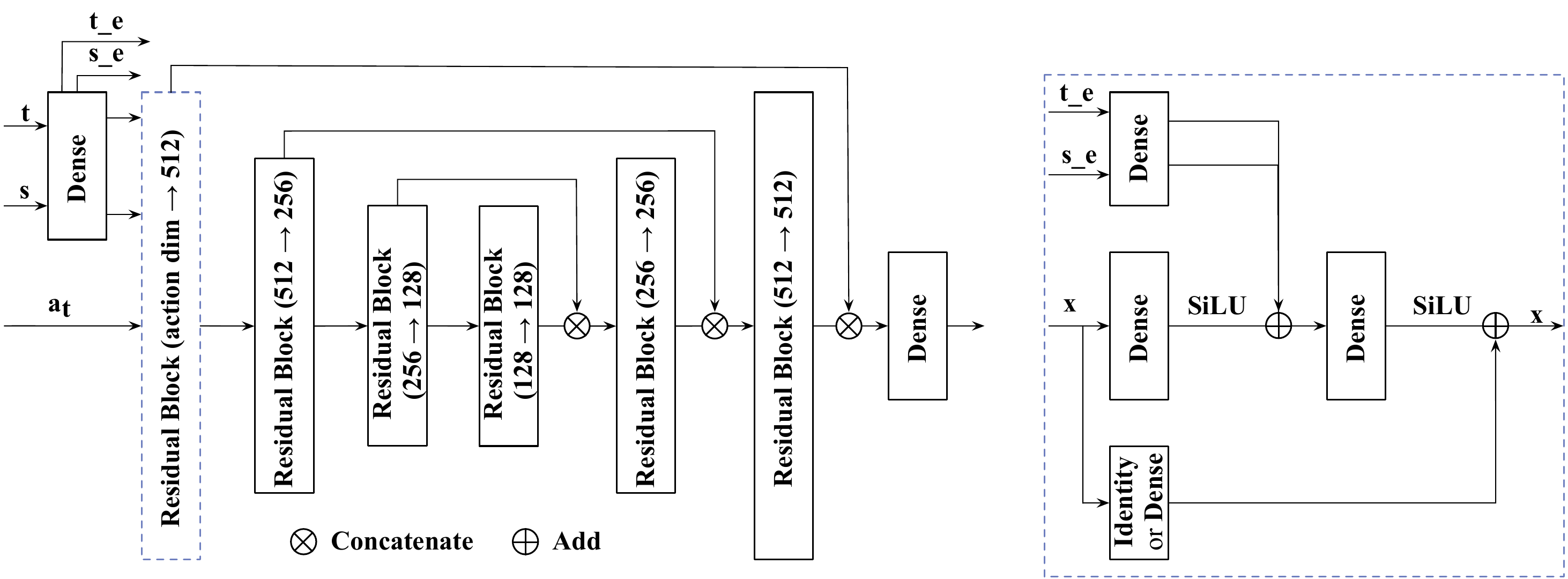}} \hspace{2em}
\subfigure[The network architecture of the state generative model.]{\includegraphics[width=0.57\textwidth]{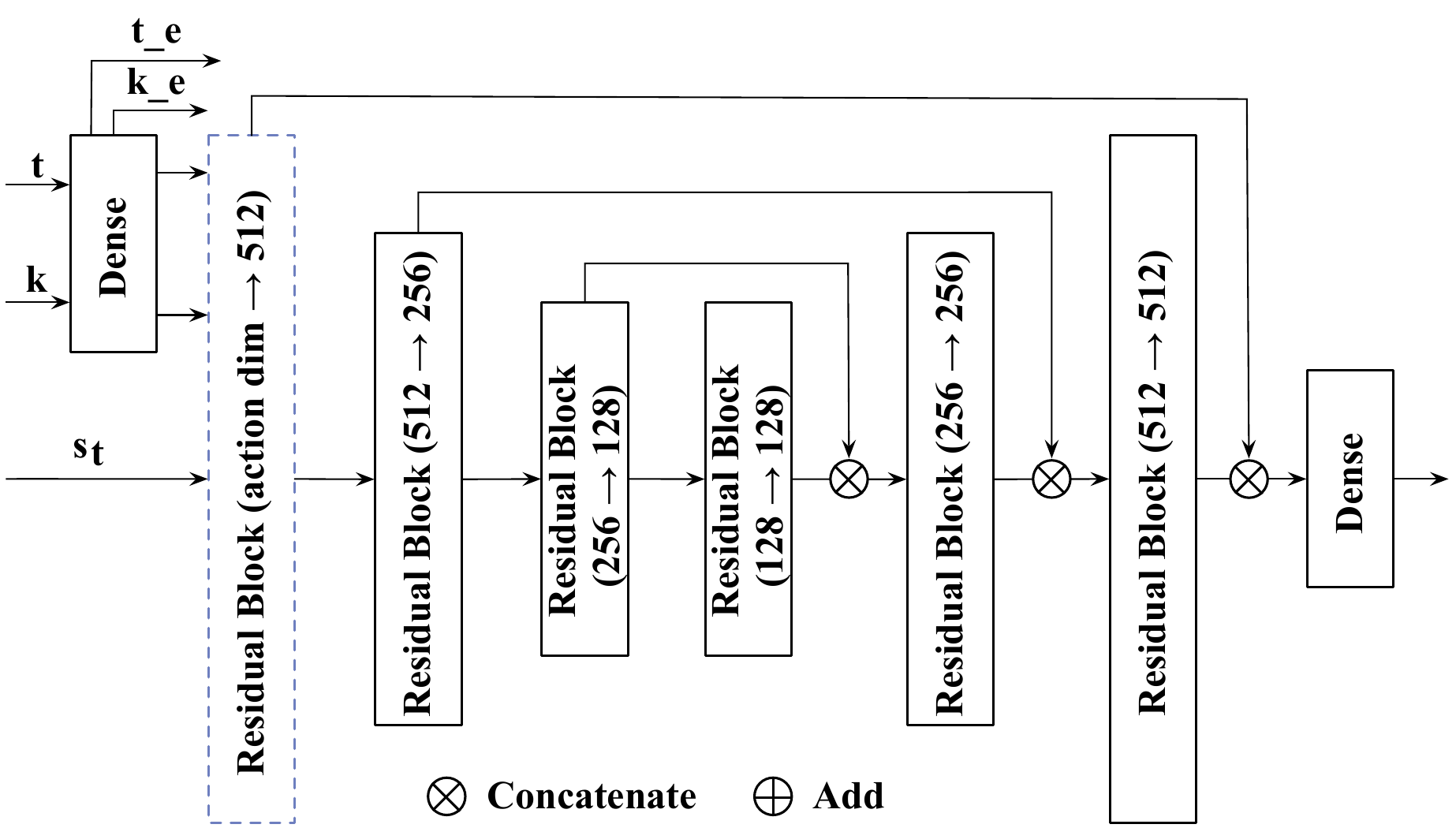}}
\caption{The network architecture of diffusion models in CuGRO.}
\label{diffusion_model}
\end{figure}

\textbf{Network Architecture.}
CuGRO includes two conditional scored-based diffusion models that estimate the score function of the behavior action distribution and the score function of the state distribution, respectively, 
and a multi-head critic model that outputs the Q-values of given state-action pairs.  
The architecture of the behavior generative model and the state generative model resembles U-Nets, but with spatial convolutions changed to simple dense connections~\citep{janner2022planning, chen2023offline}.
Please refer to Fig.~\ref{diffusion_model} for more details about the network structure. 
For the multi-head critic model, we use two hidden layers of 256 neurons with SiLU activation functions.  
We refer to the input and hidden layer as the backbone and the last output layer as the head.

\textbf{Hyperparameters.}
Table~\ref{tab3} and Table~\ref{tab4} list the main hyperparameters for diffusion models and the multi-head critic used in CuGRO, respectively.

\begin{table}[!h] \centering\setlength{\tabcolsep}{1.8mm}
\begin{minipage}[c]{0.48\textwidth}
\centering 
\caption{Hyperparameters of Diffusion models.} 
\vspace{0.5em}
\begin{tabular}{cc}
\toprule
Parameter  & Standard Configuration   \\
\midrule	
Optimizer            & Adam  \\ 
Learning rate  & $1e-4$ \\
Batch size           & $4096$  \\
Diffusion steps & $100$ \\
Epochs  & $600$ \\
$\beta_{\min }$  & $0.1$ \\
$\beta_{\max }$ & $20$ \\
\bottomrule  
\end{tabular}
\label{tab3}
\end{minipage}
\hspace{1em}
\begin{minipage}[c]{0.48\textwidth}
\centering
\caption{Hyperparameters of multi-head critic.}
\vspace{0.5em}
\begin{tabular}{cc}
\toprule
Parameter  & Standard Configuration   \\
\midrule	
Optimizer            & Adam  \\ 
Learning rate  & $1e-4$ \\
Batch size           & $4096$  \\
Epochs  & $100$ \\
Value iteration number   & $1$ \\
\bottomrule  
\end{tabular}
\label{tab4}
\end{minipage}
\end{table}

\section*{Appendix E. More Experimental Results on Superiority of Diffusion-based Replay}

\begin{figure*}[!h]\centering
 \subfigure[Swimmer-Dir]{\includegraphics[width=0.25\textwidth]{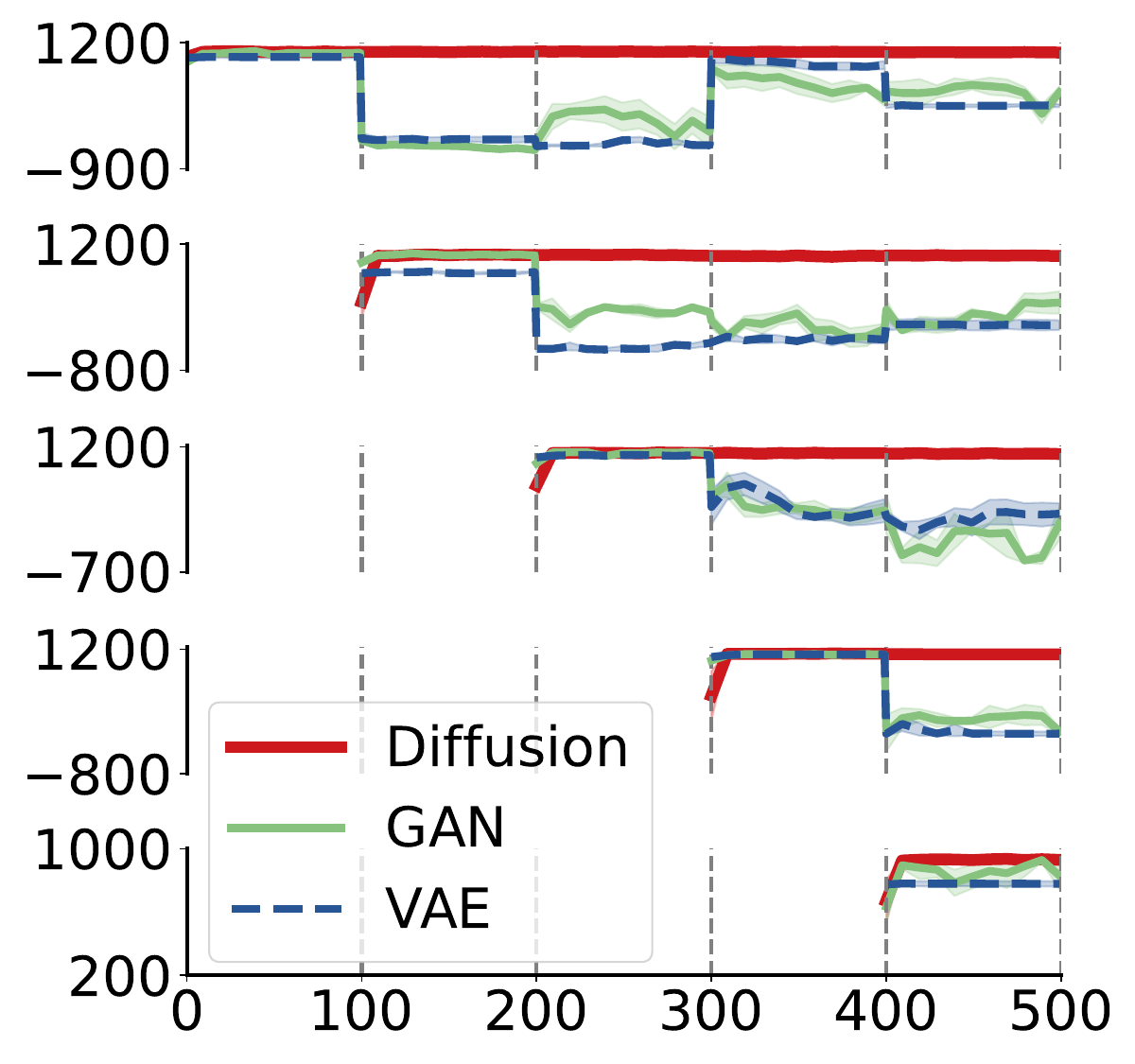}}%
 \subfigure[Walker2D-Params]{\includegraphics[width=0.25\textwidth]{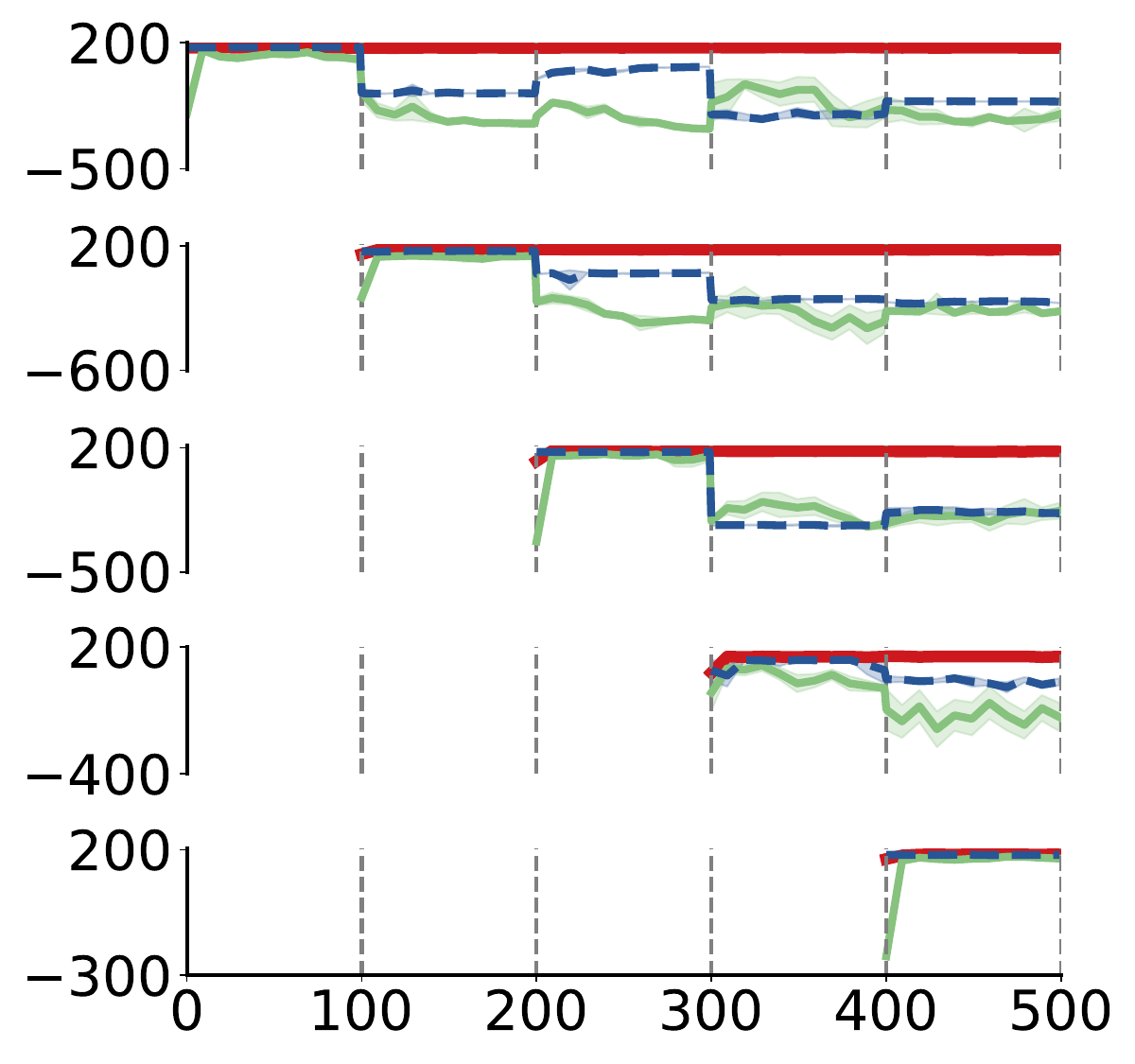}}%
 \subfigure[HalfCheetah-Vel]{\includegraphics[width=0.25\textwidth]{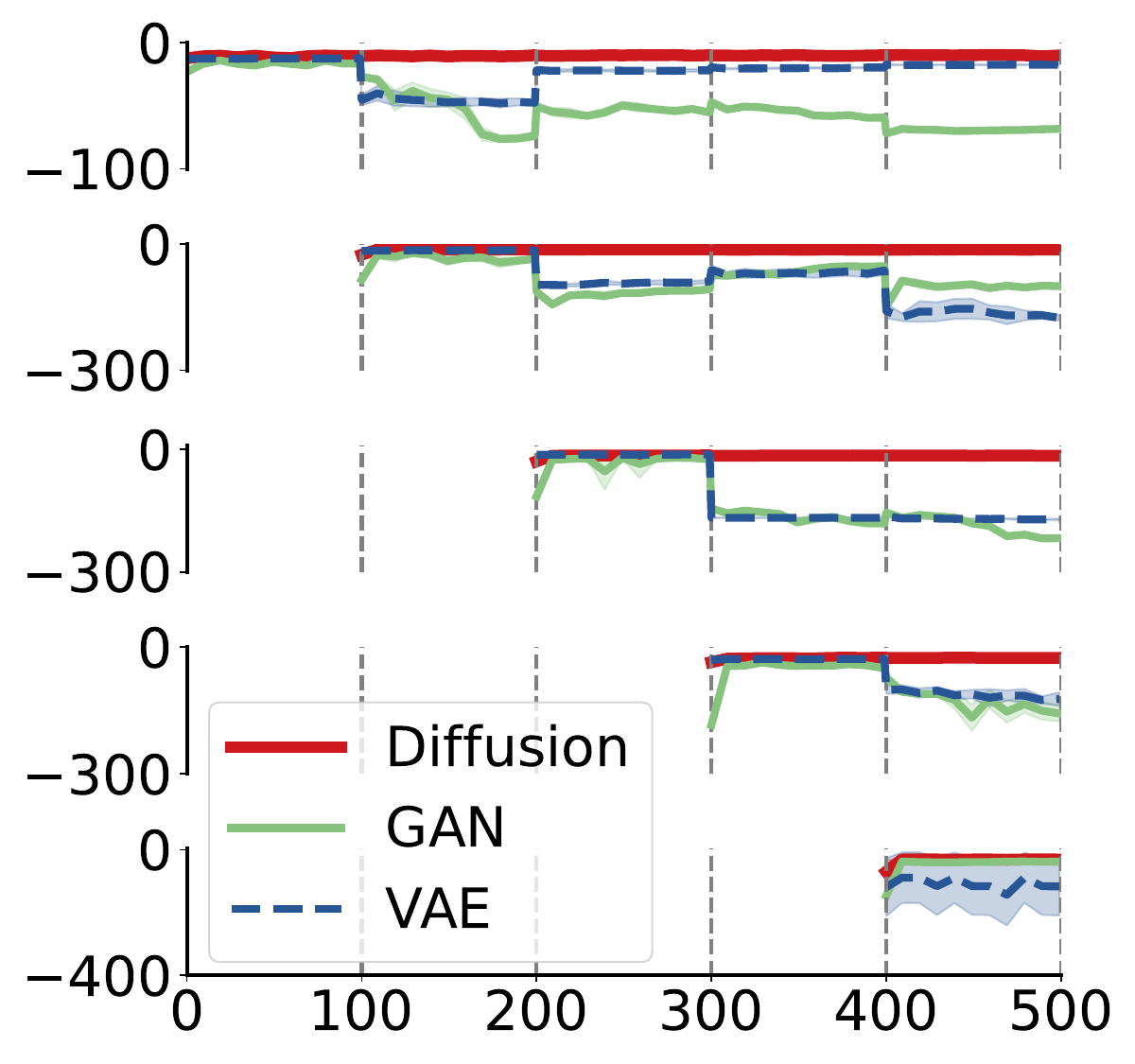}}%
 \subfigure[Meta-World]{\includegraphics[width=0.25\textwidth]{cl_models.pdf}}
 \caption{Performance of CuGRO with different generative models over  all sequential tasks on MuJoCo and Meta-World.}
 \label{fig:model_all} 
\end{figure*}

Due to limited space in the main paper, here we present the full experimental results on the Superiority of Diffusion-based Replay (Sec. 4.3) as shown in Fig.~\ref{fig:model_all} and Table~\ref{tab:model_all}.

\vspace{-1em}

\begin{table*}[!h]
\centering
\setlength{\tabcolsep}{5mm}
\caption{Final performance of CuGRO with different generative models averaged over all sequential tasks on MuJoCo and Meta-World.}
\vspace{0.5em}
\renewcommand\arraystretch{1.2}
\begin{tabular}{c|c|c|c|c}
\cmidrule[\heavyrulewidth]{1-5}
Method  & Swimmer-Dir & Walker2D-Params  & HalfCheetah-Vel   & Meta-world \\
\hline
VAE &  $  174.58 \pm 348.20 $   & $ -49.46 \pm 137.49 $  &  $ -120.94 \pm 81.85$ & $ 0.20 \pm 0.40 $  \\  

GAN  &  $ 278.43 \pm 389.07 $ & $ -107.85 \pm 150.95 $  & $ -116.21 \pm 65.37 $  & $ 0.30 \pm 0.46$ \\ 

Diffusion   &  $ \mathbf{1036.83 \pm 71.56 } $  &  $ \mathbf{173.45 \pm 10.40} $  &  $ \mathbf{-19.23 \pm 7.99}$ & $ \mathbf{1.00 \pm 0.00} $ \\ 

\cmidrule[\heavyrulewidth]{1-5}
\end{tabular}
\label{tab:model_all}
\end{table*}

\vspace{-1em}
\section*{Appendix F. More Experimental Results on Hyperparameter Analysis}
\begin{figure*}[!h]\centering
 \subfigure[Swimmer-Dir]{\includegraphics[width=0.25\textwidth]{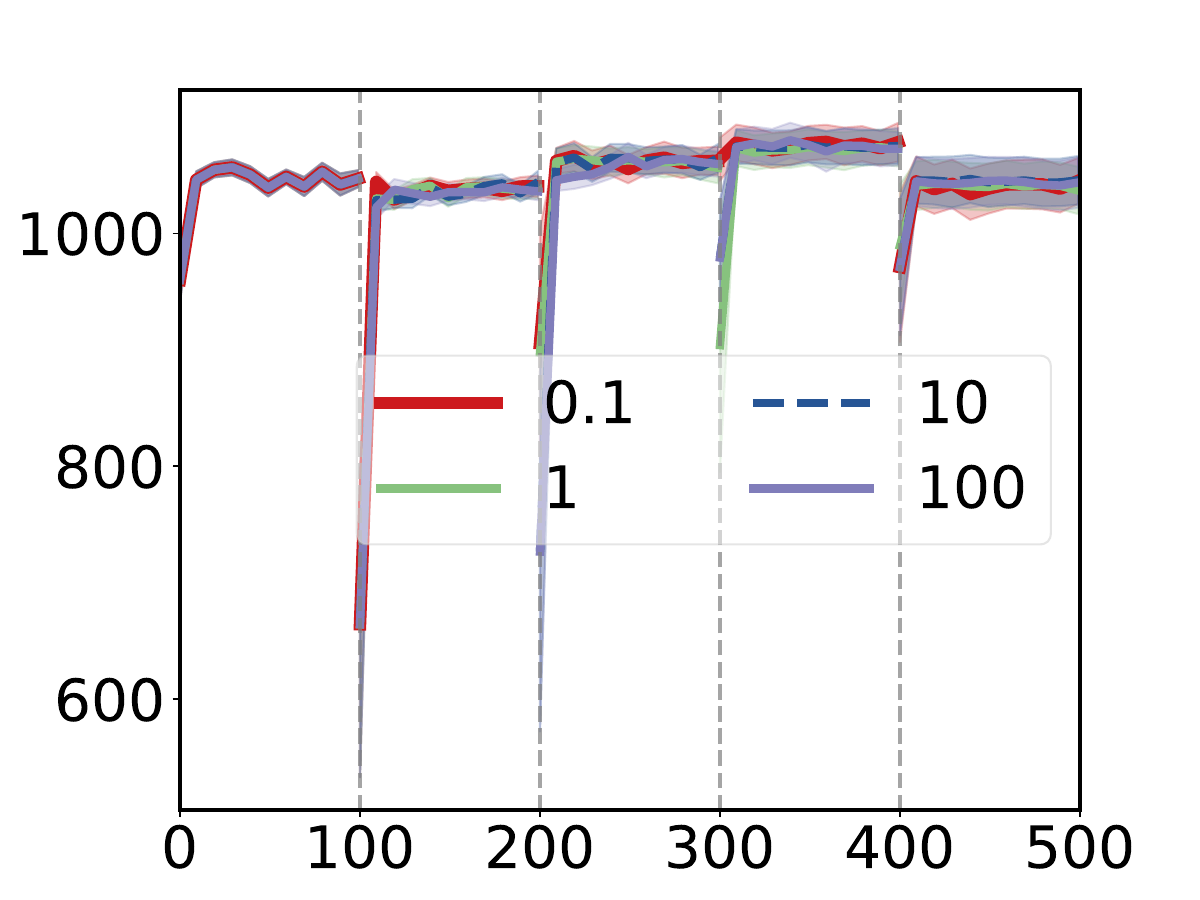}}%
 \subfigure[Walker2D-Params]{\includegraphics[width=0.25\textwidth]{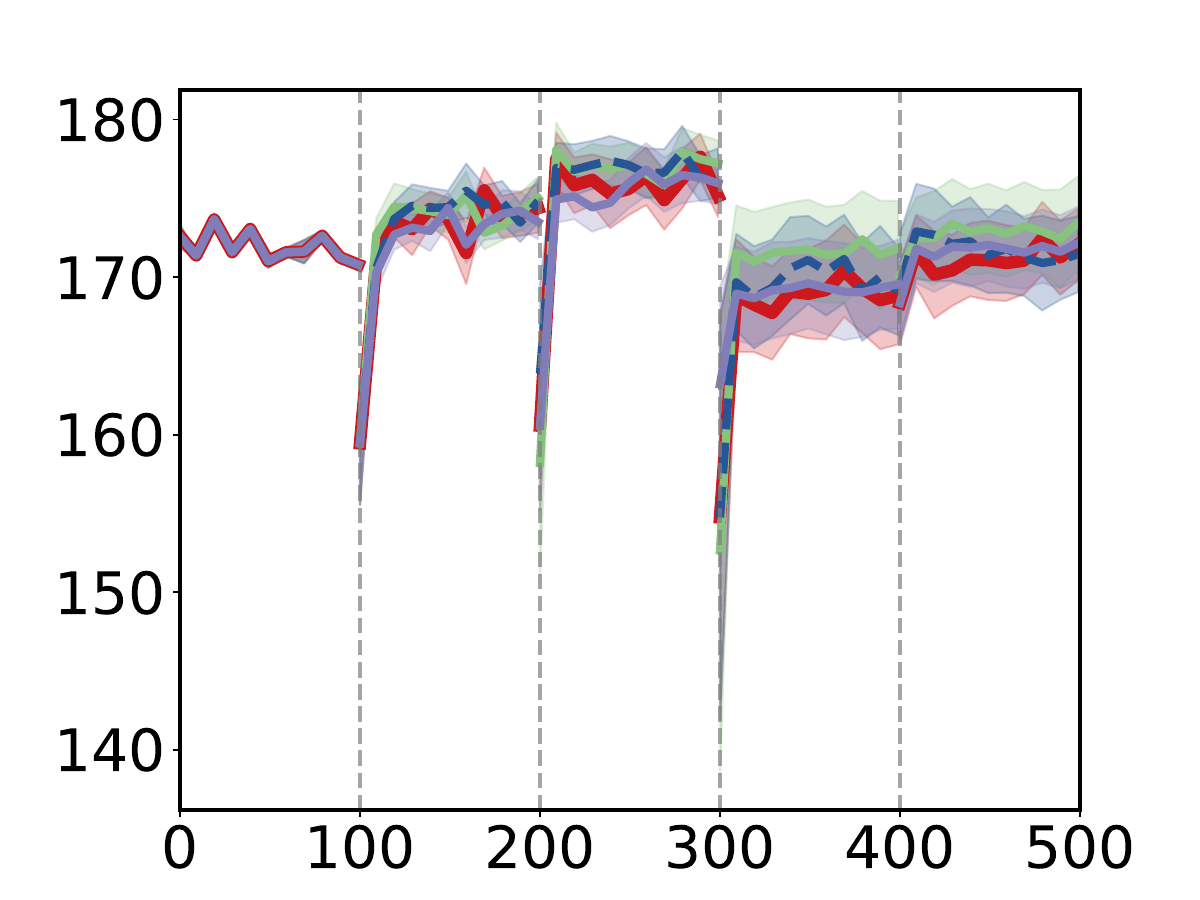}}%
 \subfigure[HalfCheetah-Vel]{\includegraphics[width=0.25\textwidth]{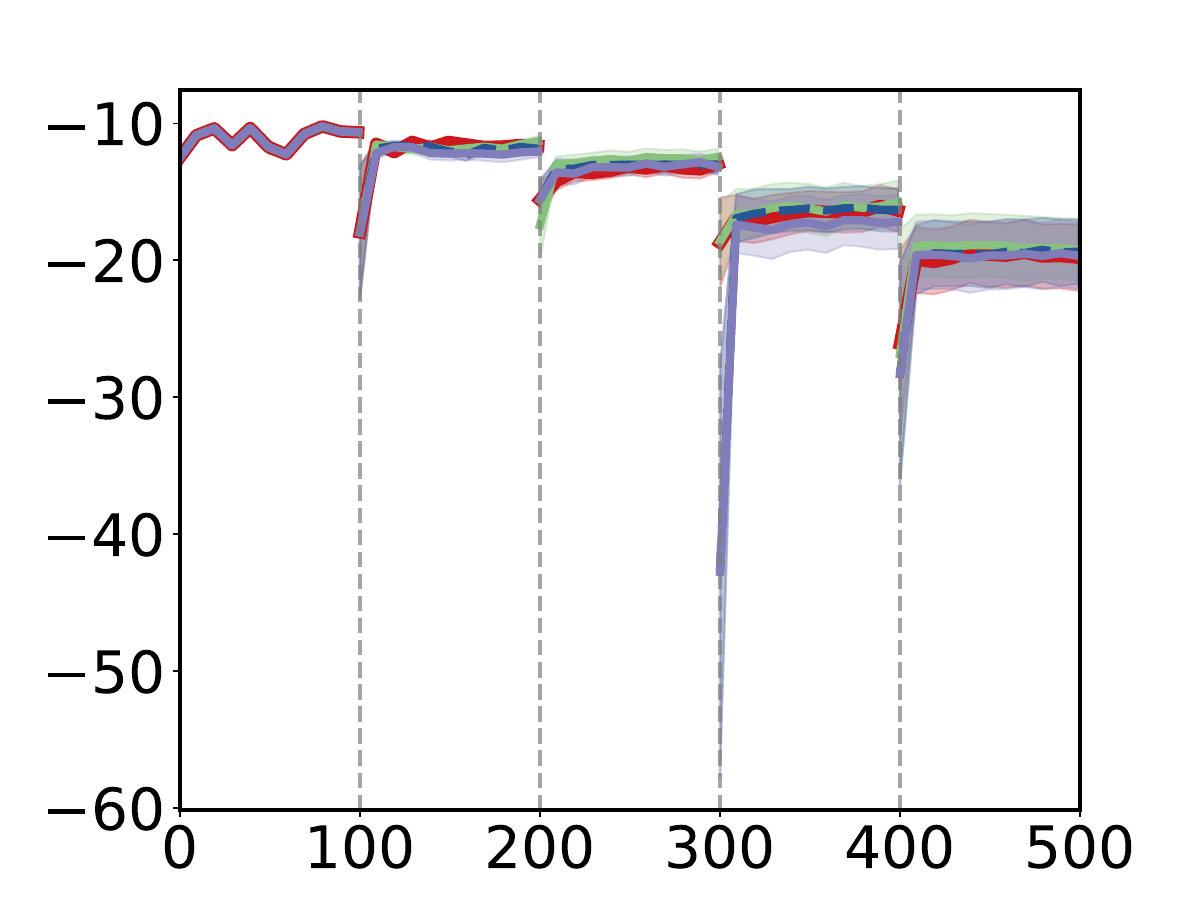}}%
 \subfigure[MetaWorld]{\includegraphics[width=0.25\textwidth]{cl_l.pdf}}
 \caption{Performance of CuGRO with varying coefficients $\lambda$ over cumulative tasks on MuJoCo and Meta-World.}
 \label{fig:hyper} 
\end{figure*}

Due to limited space in the main paper, here we present the full experimental results on the Hyperparameter Analysis (Sec. 4.4) as shown in Fig.~\ref{fig:hyper} and Table~\ref{tab:all_hyper}.

\vspace{-1em}

\begin{table*}[!h]
\centering
\setlength{\tabcolsep}{5mm}
\caption{Final performance of CuGRO with varying coefficients $\lambda$ averaged  over cumulative tasks on MuJoCo and Meta-World.}
\vspace{0.5em}
\renewcommand\arraystretch{1.2}
\begin{tabular}{c|c|c|c|c}
\cmidrule[\heavyrulewidth]{1-5}
$\lambda$  & Swimmer-Dir & Walker2D-Params  & HalfCheetah-Vel   & Meta-World \\
\hline
0.1   & $ 1045.09 \pm 69.33 $   &  $ 172.07 \pm 8.01 $  & $ -19.80 \pm 8.53  $  &  $ 1.00 \pm 0.00 $  \\ 

1   &  $ 1036.83 \pm  71.56 $    &  $ \mathbf{173.45 \pm 10.40 }  $  &  $ \mathbf{-19.23 \pm  7.99 }$ & $ 1.00 \pm 0.00 $ \\ 

10 &  $ \mathbf{1045.61 \pm 73.56} $   & $ 171.47 \pm 8.52 $  &  $  -19.37 \pm 8.28  $ & $ 1.00 \pm 0.00 $  \\ 

100 &  $ 1043.70 \pm 76.77 $  & $   172.17 \pm 8.12  $  &  $  -19.66 \pm 8.58 $  &  $ 1.00 \pm 0.00 $ \\ 
\cmidrule[\heavyrulewidth]{1-5}
\end{tabular}
\label{tab:all_hyper}
\end{table*}

\section*{Appendix G. More Experimental Results on datasets of different qualities}
\begin{figure*}[!h]\centering
 \subfigure[Swimmer-Dir]{\includegraphics[width=0.25\textwidth]{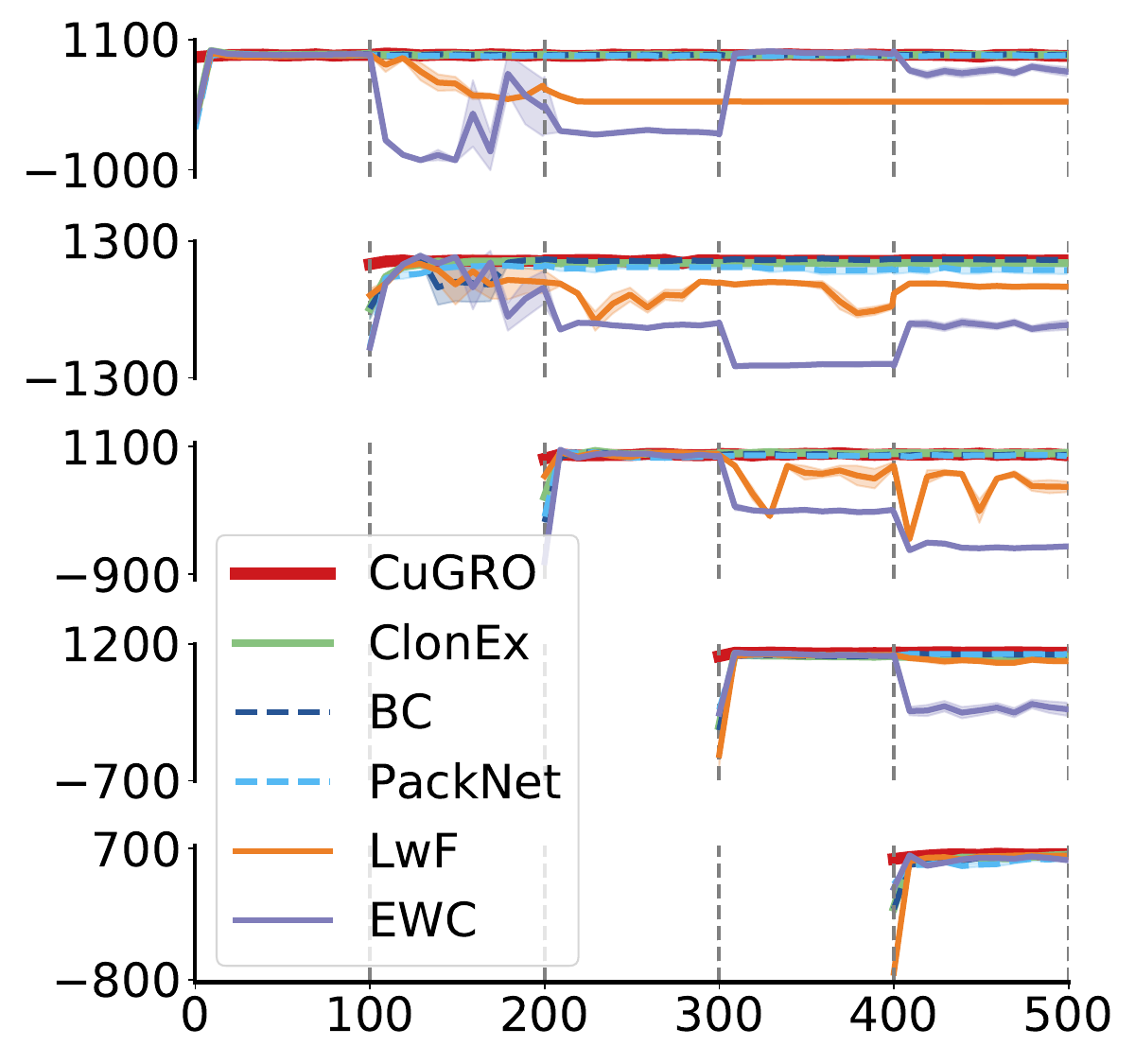}}%
 \subfigure[Walker2D-Params]{\includegraphics[width=0.25\textwidth]{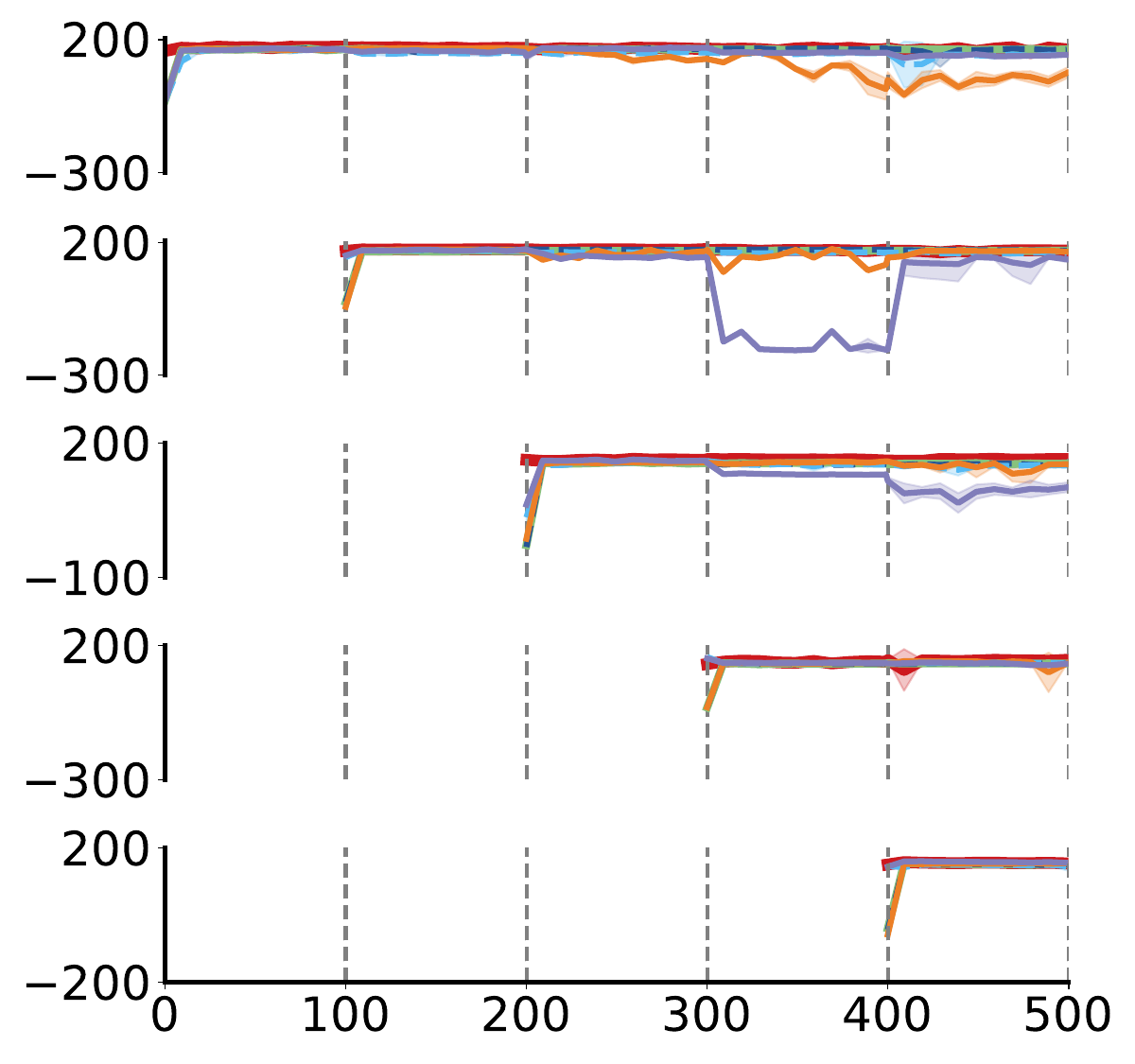}}%
 \subfigure[HalfCheetah-Vel]{\includegraphics[width=0.25\textwidth]{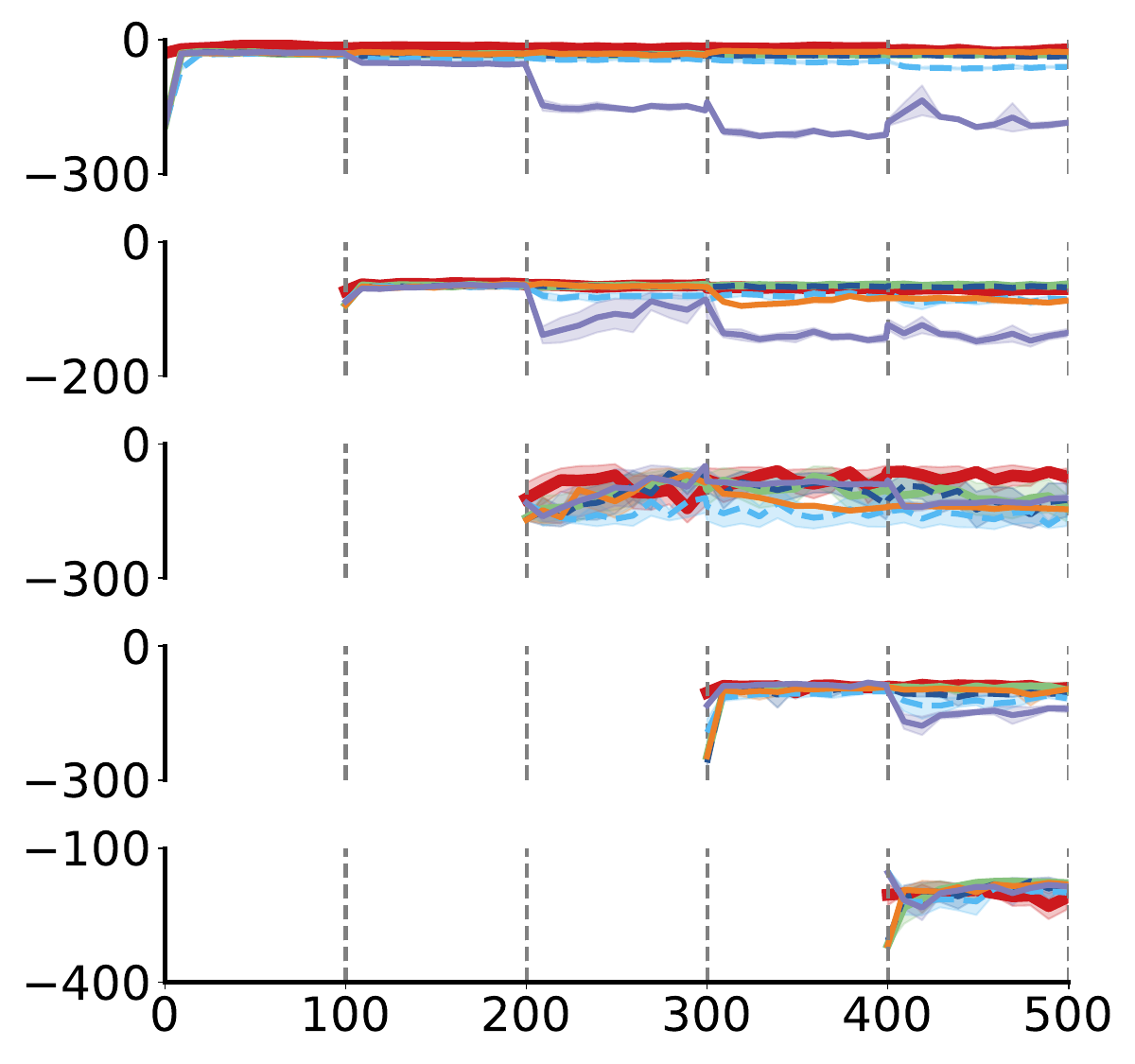}}%
\subfigure[Meta-World]{\includegraphics[width=0.25\textwidth]{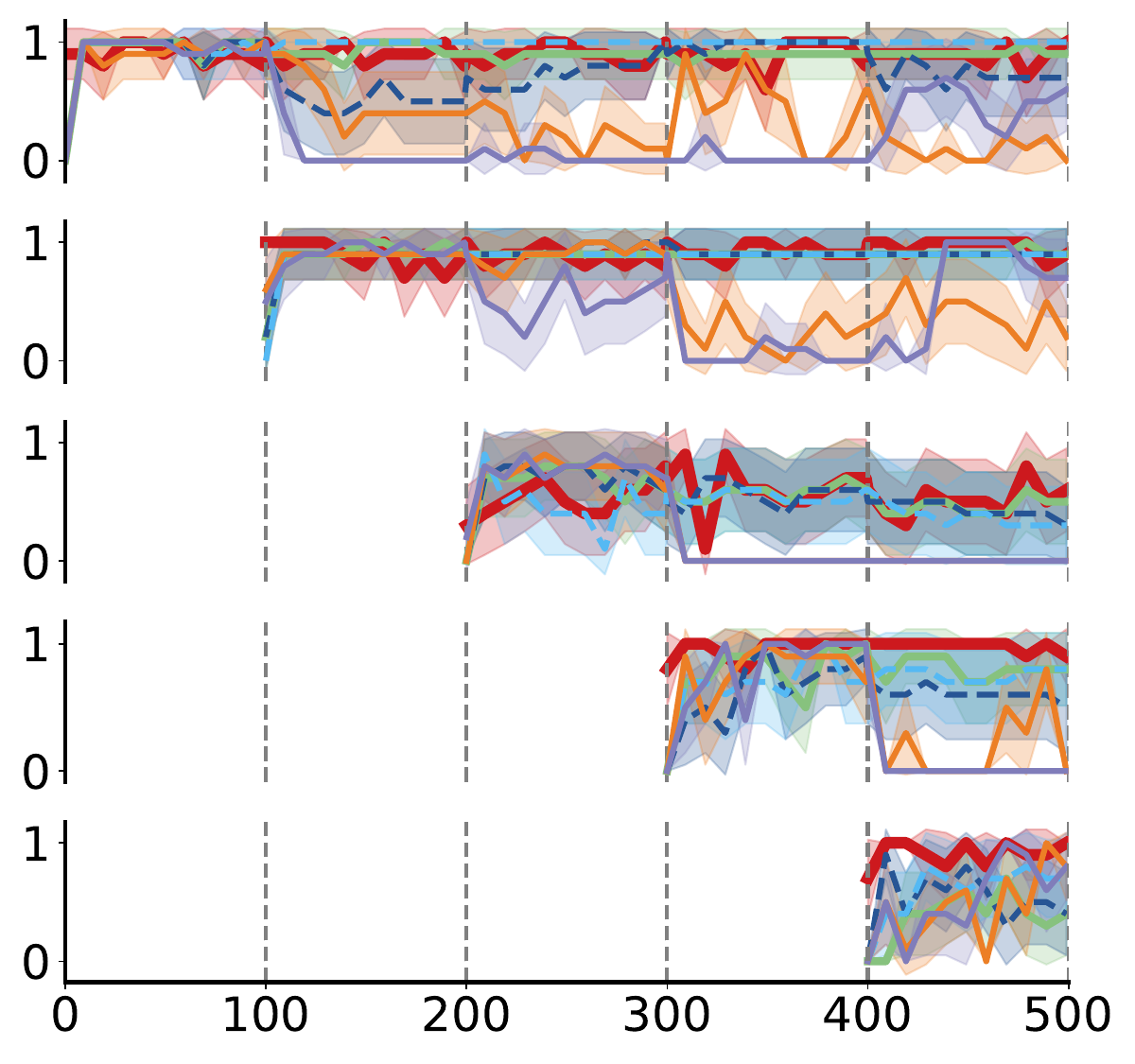}}
 \caption{Performance of CuGRO and baselines on cumulative tasks during sequential training evaluated on MuJoCo and Meta-World on medium dataset  }
\label{fig:medium} 
\end{figure*}

\begin{figure*}[!h]\centering
 \subfigure[Swimmer-Dir]{\includegraphics[width=0.25\textwidth]{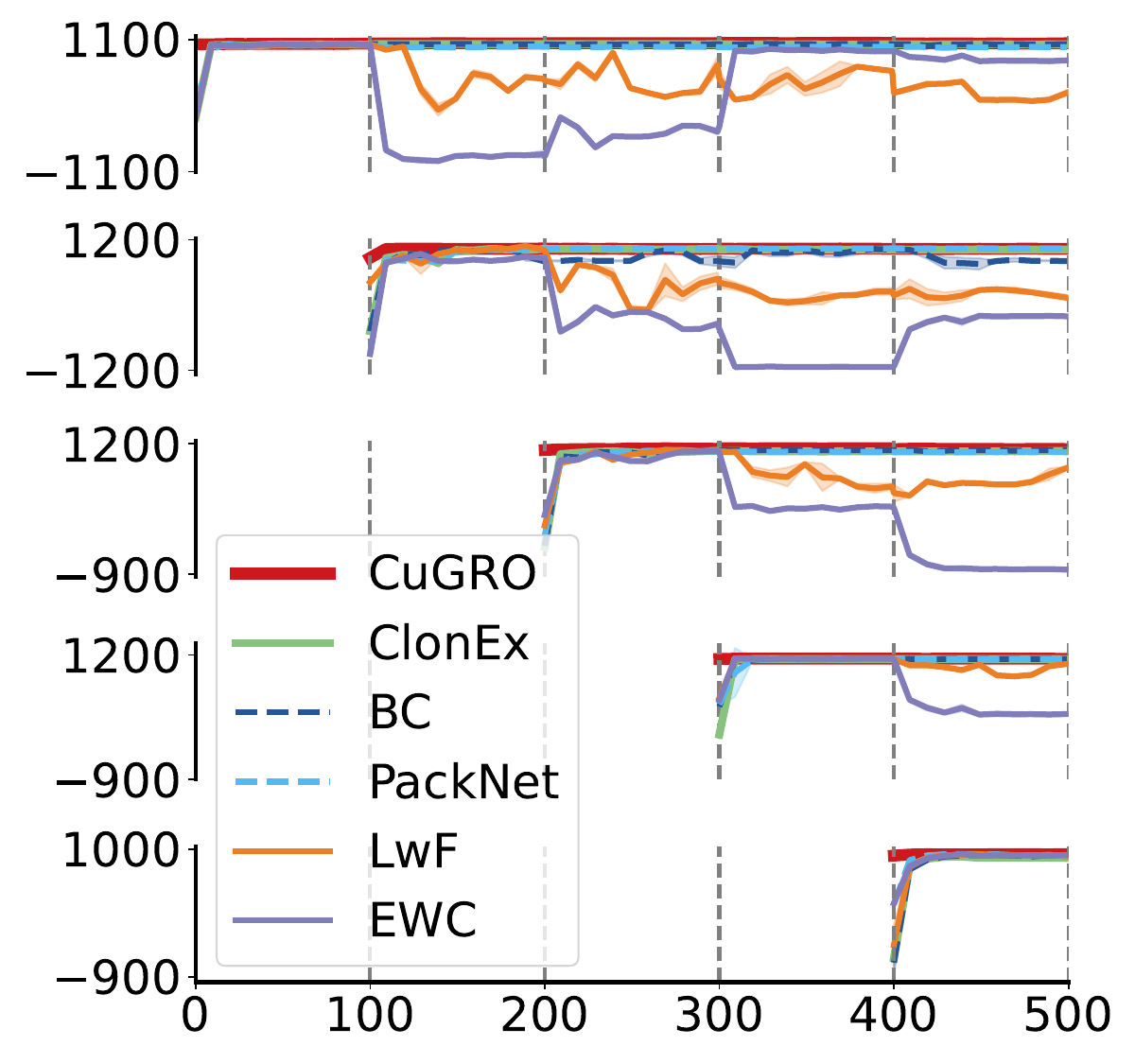}}%
 \subfigure[Walker2D-Params]{\includegraphics[width=0.25\textwidth]{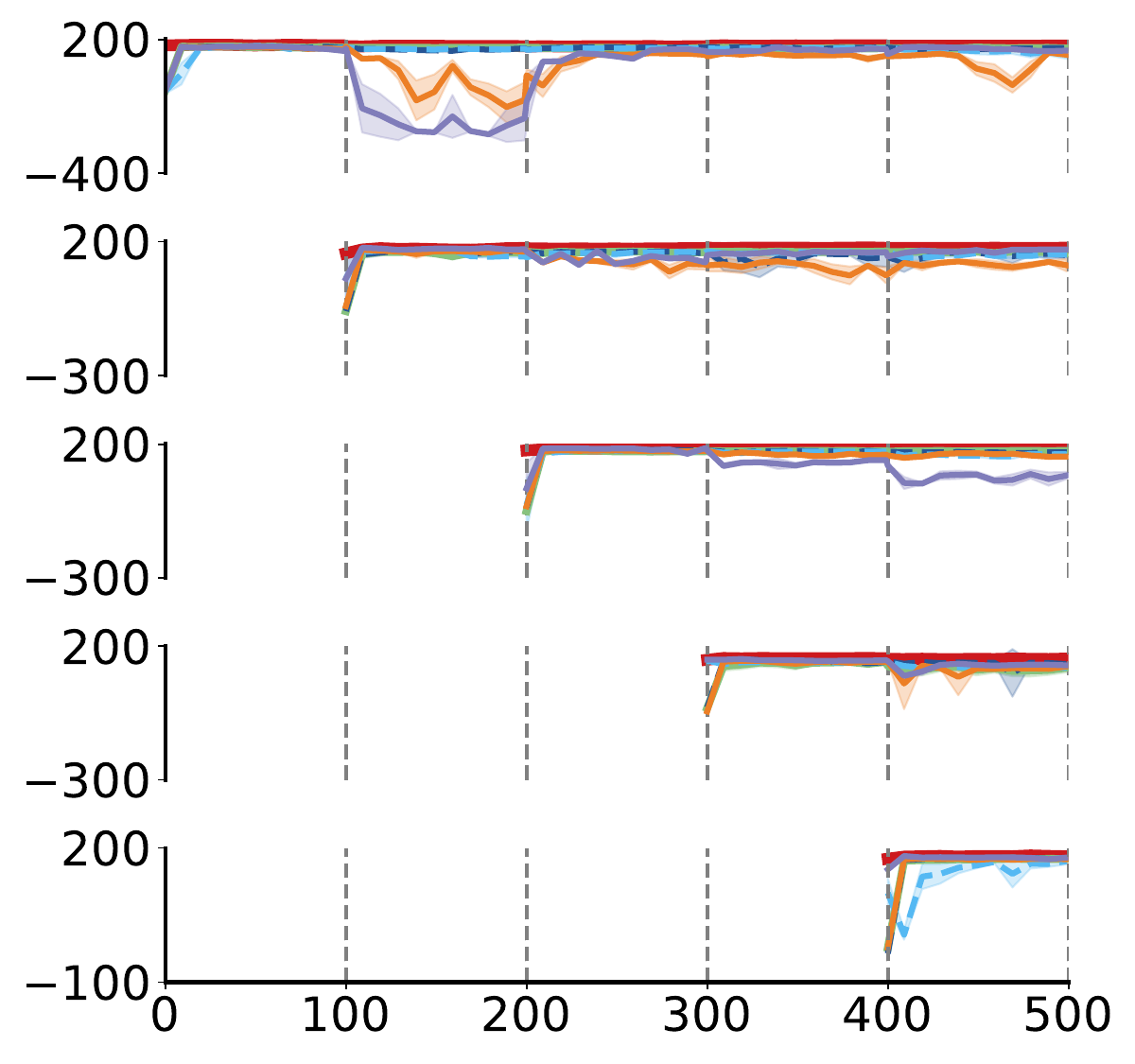}}%
 \subfigure[HalfCheetah-Vel]{\includegraphics[width=0.25\textwidth]{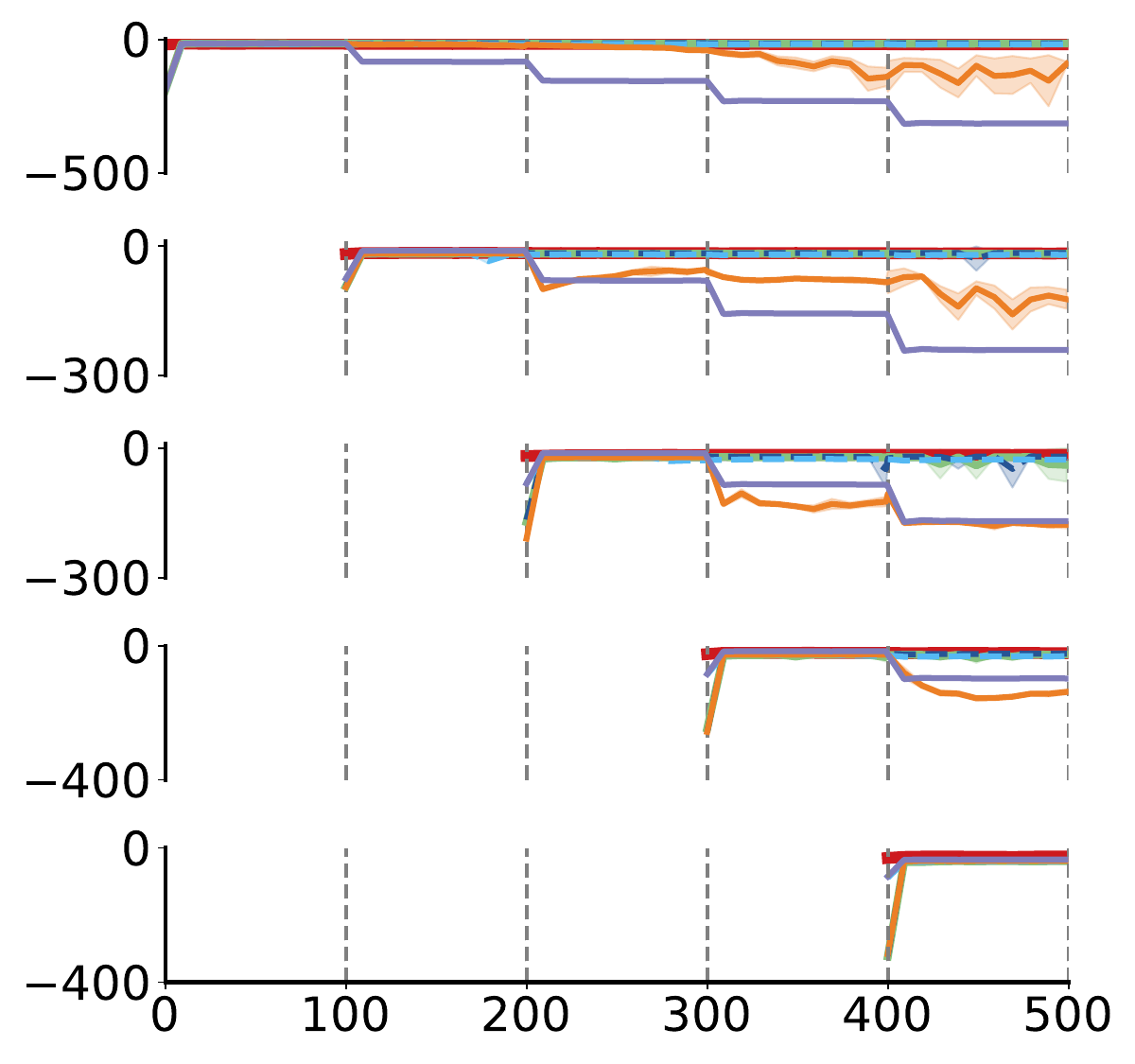}}%
\subfigure[Meta-World]{\includegraphics[width=0.25\textwidth]{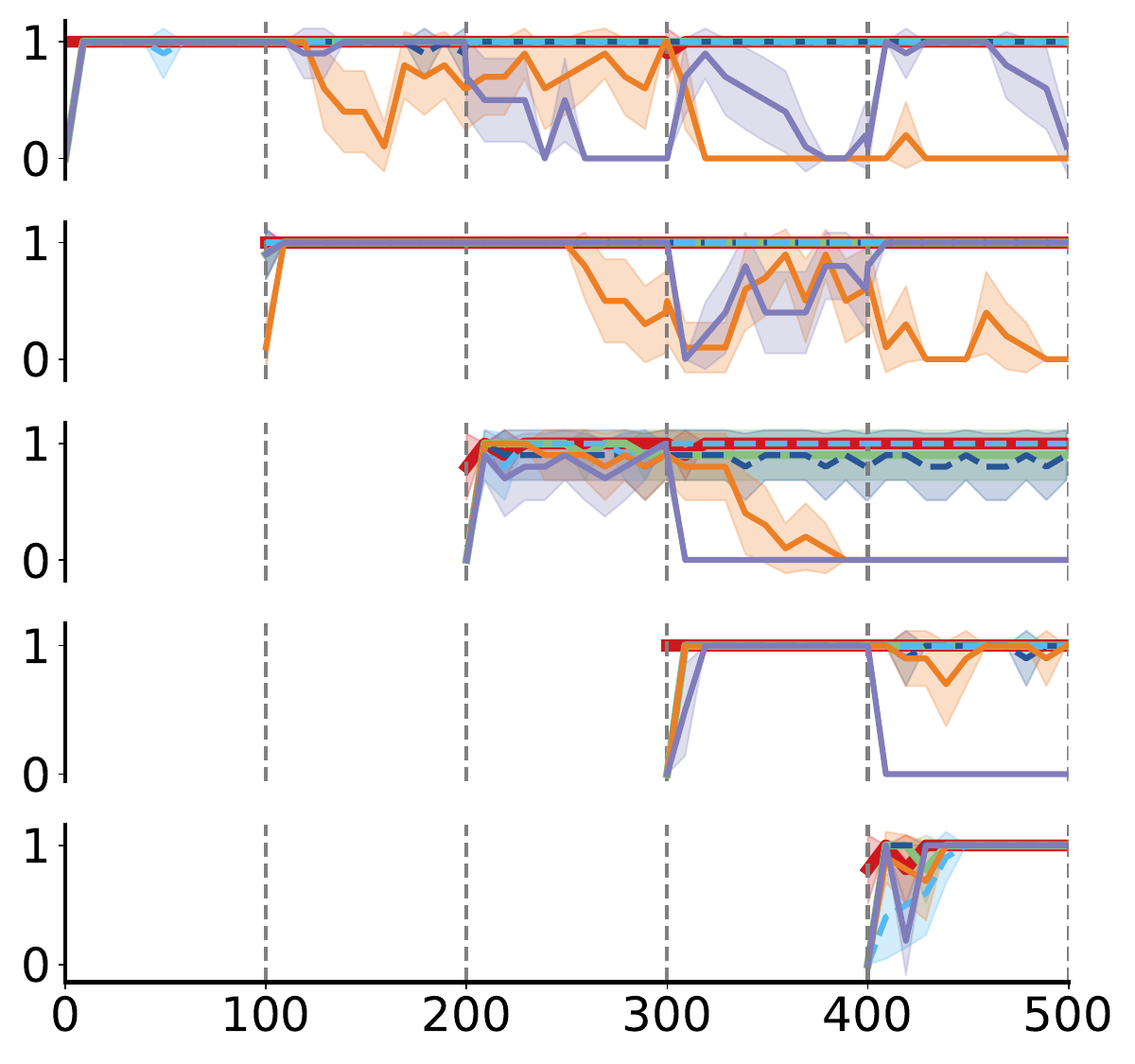}}
 \caption{Performance of CuGRO and baselines on cumulative tasks during sequential training evaluated on MuJoCo and Meta-World on expert dataset. }
\label{fig:expert} 
\end{figure*}

Due to limited space in the main paper, we present the experimental results on datasets of expert and medium qualities (Sec. 4.1), as shown in Figures \ref{fig:medium} and \ref{fig:expert}, and Tables \ref{tab:medium} and \ref{tab:expert}. 

\vspace{-1em}

\begin{table*}[!h]
\centering
\setlength{\tabcolsep}{5mm}
\caption{Final performance of CuGRO and baselines averaged over all sequential tasks evaluated on MuJoCo and Meta-World on medium dataset .}
\renewcommand\arraystretch{1}
\begin{tabular}{c|c|c|c|c}
\cmidrule[\heavyrulewidth]{1-5}
Method  & Swimmer-Dir  & Walker2D-Params  & HalfCheetah-Vel   & Meta-World \\
\hline
EWC   & $138.36 \pm 448.87 $  &  $ 133.56 \pm 21.01 $    & $ -153.93 \pm 28.33 $ & $ 0.42 \pm 0.49 $  \\

LwF   & $ 516.60 \pm  287.18 $   &  $ 135.96 \pm 35.14 $    & $ -107.91 \pm  52.80 $ & $0.20 \pm 0.40 $ \\

PackNet  & $839.73 \pm 169.09$  &  $ 145.93 \pm 13.06 $    & $ -123.46 \pm 54.30 $ & $ 0.74 \pm 0.44 $\\
 
BC &  $ 887.98 \pm 152.43 $   & $ 153.59 \pm 12.41 $  &  $-103.11 \pm 54.45$ & $ 0.56 \pm 0.50  $ \\

ClonEx  & $877.37 \pm 145.72$   & $ 154.31 \pm  12.70 $  &  $-103.67 \pm 56.83$ & $ 0.7 \pm 0.46 $ \\

CuGRO & $\mathbf{  889.45 \pm 150.82} $ &  $ \mathbf{159.69 \pm 9.52} $  &  $ \mathbf{-94.07 \pm 66.61} $ & $ \mathbf{ 0.88 \pm 0.32 }$ \\ 
\cmidrule[\heavyrulewidth]{1-5}
\end{tabular}
\label{tab:medium}
\end{table*}
\vspace{-1em}

\begin{table*}[!h]
\centering
\setlength{\tabcolsep}{5mm}
\caption{Final performance of CuGRO and baselines averaged over all sequential tasks evaluated on MuJoCo and Meta-World on expert dataset .}
\renewcommand\arraystretch{1}
\begin{tabular}{c|c|c|c|c}
\cmidrule[\heavyrulewidth]{1-5}
Method  & Swimmer-Dir  & Walker2D-Params  & HalfCheetah-Vel   & Meta-World \\
\hline
EWC   & $167.73 \pm 637.52 $  &  $ 142.27 \pm 35.48 $    & $ -171.34 \pm 99.36 $ & $  0.42 \pm 0.49 $  \\

LwF   & $620.26 \pm 375.77$   &  $ 139.45 \pm 27.55 $    & $ -113.56 \pm 49.42 $ & $ 0.4  \pm 0.49 $ \\

PackNet   & $ 1022.88 \pm  80.40 $  &  $ 148.50 \pm 24.85 $    & $ -27.14 \pm  8.60 $ & $ 1.0 \pm 0.0 $\\

BC &  $  991.94 \pm 123.51 $   & $ 158.42 \pm 17.56 $  &  $ -22.07 \pm 8.43 $ & $ 0.98\pm 0.14  $ \\

ClonEx &  $ 1025.18 \pm 91.00$   & $  157.34 \pm 23.69 $  &  $ -27.43 \pm 26.36  $ & $ 0.98\pm 0.14  $ \\

CuGRO & $ \mathbf{1049.67 \pm 78.31} $ &  $ \mathbf{173.50 \pm 10.66} $  &  $ \mathbf{-19.54 \pm 3.84} $ & $ \mathbf{1.0 \pm 0.0 }$ \\ 
\cmidrule[\heavyrulewidth]{1-5}
\end{tabular}
\label{tab:expert}
\end{table*}

\section*{Appendix H. Limitations and broader impacts}
As the first attempt, CuGRO opens up avenues for leveraging powerful diffusion models to solve sequential offline RL tasks.
The demonstrated improvements in forward transfer and reduced catastrophic forgetting highlight the potential impact of this continual learning framework that does not require accessing real samples of previous tasks.
This is promising for a wide range of applications, especially scenarios concerning data privacy and safety, like federated learning, healthcare, and finance.
\end{document}